\documentclass[a4paper,twoside]{article}
\usepackage{epsfig}
\usepackage{subfigure}
\usepackage{calc}
\usepackage{amssymb}
\usepackage{amstext}
\usepackage{amsmath}
\usepackage{multicol}
\usepackage{pslatex}
\usepackage{apalike}
\usepackage{fancyhdr}
\usepackage{insticc}
\usepackage{bm}
\usepackage[small]{caption}
\begin{document}
\onecolumn \maketitle \normalsize \vfill
%%%%%%%%%%%%%%%%%%%%%%%%%%%%%%%%%%%%%%%
\section{Introduction}
\label{sec:introduction}
%%%%%%%%%%%%%%%%%%%%%%%%%%%%%%%%%%%%%%%%
Genetic Algorithm (GA) is a heuristics to find the best possible 
solution for combinatorial optimization problems and 
it is based on several relevant operators 
such as selection, crossover and mutation on the 
gene configurations (strings) leading to transition from one state to the others.
This probabilistic algorithm was firstly introduced 
in the book {\it Adaptation in Natural and Artificial Systems}  
by John Holland in 1975 \cite{Holland} and now 
it has been widely used in various research fields and 
established as one of the effective algorithms to 
find the solution within reasonable computational time \cite{Goldberg}. 

However, it does not mean that 
the GA can be automatically applied to 
any problem and can find the candidates of the solutions 
immediately. 
We should choose the suitable information representation of 
each gene (member) in population (ensemble) for a given problem and 
the parameters which control the operations should be set to 
the optimal values. 
Basically, these operations 
except for the selection, which 
is dependent on 
the fitness,  
are defined as procedures to generate 
the states randomly and the operations 
do not always confirm to increase the fitness value. 
To make matter worse, 
there are few mathematical justification 
for the GA to make the 
system to convergence to one of  the best possible solution. 

From this fact in mind, 
in this paper, 
in order to figure out 
the statistical properties of GA from 
the view point of thermodynamics, 
we introduce a learning algorithm 
of Gibbs distributions 
from training sets which are 
gene configurations generated by GA. 
A procedure of 
average-case performance evaluation for genetic algorithms 
is examined. 
The learning algorithm is 
constructed by means of 
minimization of the 
Kullback-Leibler information between a 
parametric Gibbs distribution and the empirical distribution 
of gene configurations. 
The formulation is applied to 
the solvable probabilistic models 
having multi-valley energy landscapes, 
namely, the spin glass chain \cite{Li,Chen} and 
the Sherrington-Kirkpatrick model \cite{SK} in statistical physics. 
By using computer simulations, we discuss the asymptotic 
behaviour of the effective 
temperature scheduling and the residual energy induced by the GA dynamics. 
We also reveal the operator-dependence of the behaviour. 

As well-known, a study focusing on 
the distribution of gene configurations in GA 
itself is not a bran-new approach. 
Actually, the so-called {\it Estimation of Distribution Algorithm (EDA)} 
\cite{Baluja,Pelikan1,Pelikan2,Pelikan3,Shapiro,Shapiro2,Shapiro3} 
is a well-known and established approach to find the best possible solution 
by estimating the distribution of 
gene configuration 
during the GA dynamics and 
one can use the distribution to 
produce the genes in the next generation. 
In fact, 
a lots of such studies have been done 
for various problems. 

For instance, 
Prugel-Bennett and Shapiro \cite{Bennett1,Bennett2} 
evaluated the time evolution of 
the cummulants of distributions  
and discussed the statistical properties of 
GA from the dynamical point of view. 
Suzuki \cite{Suzuki,Suzuki2,Suzuki3} represented the relationship between 
the gene configurations by graphical models and 
estimated the joint probability or 
the marginal probability of the genes 
by making use of Belief propagation on the graphical models.   

Nevertheless there exist such extensive studies, 
in the present study, 
we choose a Gibbs distribution 
which is specified by a single 
parameter, namely, temperature $T$ and 
learns the distribution 
(the effective temperature) 
from the gene configurations 
produced by GA. 
Thus, 
we attempt to figure out 
the average-case performance of GA 
from the view point of 
temperature scheduling in simulated annealing. 
Moreover, the evaluation of 
average-case performance 
is partially carried out analytically by 
choosing the energy function 
of solvable spin glass models. 
These points are remarkable distinctions of 
our approach in the present paper.  

This paper is organized as follows.
In the next section, we mention the relationship between 
the GA and simulated annealing from the view point 
of the distribution of ensembles (population) on Markovian process. 
In section 3, we explain our formulation and 
tools to investigate the average-case performance 
of the GA. We construct the Boltzmann-machine-type 
learning equation 
via the minimization of Kullback-Leibler information 
between the empirical distribution 
of GA and a Gibbs distribution with respect to 
the effective temperature. 
The validity of a Gibbs form of the distribution is confirmed 
by the so-called Holland's condition. 
The learning equation is rewritten 
in terms of 
optimization of 
the energy function 
for Ising spin systems. 
The concept of average-case performance 
is mentioned, 
namely, the so-called self-averaging properties for 
physical quantities and the replica method to 
carry out the average are introduced.  
In the next section 4, we introduce our benchmark test 
problem, namely, the combinatorial optimization 
problem having the energy function 
of the so-called spin glasses. 
The mathematically tractable spin glasses, 
namely, spin glass chain and 
the Sherrington-Kirkpatrick model 
are introduced and their statistical properties are revealed.  
In section 5, we explain 
the set-up of our numerical 
experiments and 
the results are reported in the next section 6. 
The last section is devoted to concluding remarks. 
%%%%%%%%%%%%%%%%%%%%%%%%%%%%%%%%%%%%%%%%%%%%%%%
\section{GA and SA}
%%%%%%%%%%%%%%%%%%%%%%%%%%%%%%%%%%%%%%%%%%%%%%%%
As we mentioned, in this paper, we consider the statistical properties of 
GA from the view point of thermodynamics. 
In simple GA, we define each 
gene configuration (member) by 
a string of binary variables with length $N$, 
that is, 
$\bm{s} 
=(s_{1},s_{2},\cdots,s_{N}), s_{i} \in \{-1,+1\}$, 
and we attempt to make each configuration in 
ensemble with size $M$ to the state 
which gives a minimum of 
the energy function $H(\bm{s})$, say, 
$\bm{s}_{*}$．
The problem is systematically solved by GA if 
the system evolves 
according to a Markovian process 
and the gene distribution $P_{GA}^{(t)}(\bm{s})$ 
at time (generation) 
$t$ might converge as 
$P_{GA}^{(t)}(\bm{s}) \to  
P_{GA}^{(\infty)}(\bm{s})$ and 
we have 
%%%%%%
\begin{equation} 
P_{GA}^{(\infty)}(\bm{s}) = 
\delta (\bm{s}-\bm{s}_{*})
=\prod_{i=1}^{N}
\delta (s_{i}-s_{i*}). 
\end{equation}
%%%%%%
On the other hand, 
one of the effective heuristics 
which is well-known as 
{\it Simulated Annealing (SA)} \cite{Kirkpatrick,Geman} is achieved by 
inhomogeneous Markovian process. 
The process 
is realized by  Markov chain Monte Carlo method (MCMC) 
which leads to an equilibrium Gibbs distribution 
at temperature $T=\beta^{-1}$ (from now on, 
the $\beta$ is referred to
 as `inverse temperature'), namely, 
%%%%%%
\begin{equation}
P_{B}^{(t)}(\bm{s})  =  
\frac{{\rm e}^{-\beta^{(t)} H(\bm{s})}}
{Z},\,\,\,\,
Z=
\sum_{\bm{s}}
{\rm e}^{-\beta^{(t)} H(\bm{s})}.
\end{equation}
%%%%%%
In SA, the temperature is 
scheduled very slowly in time as 
$\beta^{(\infty)} \to \infty$ 
($T^{(\infty)} \to 0$), and then, we can solve the problem as 
%%%%%%
%%%%%%
\begin{equation}
P_{B}^{(\infty)}(\bm{s}) = 
\delta (\bm{s}-\bm{s}_{*})
=\prod_{i=1}^{N}
\delta (s_{i}-s_{i*}). 
\label{eq:SA}
\end{equation}
%%%%%%
Therefore, 
both the GA and the SA 
share a concept to 
make the distribution convergence to 
a single (or several) delta-peak(s) at the solution(s). 
However, in general, the Markovian (dynamical) process of 
GA is very hard to 
treat mathematically due to 
the global transition between the states 
by the crossover or, especially,  the mutation 
operator, whereas
the SA causes only local transitions  
between the states. 
From the view point of EDA, 
the dynamics of GA should lead to an empirical distribution 
of states.  
As we shall mention later on, 
the distribution is more likely to be a Gibbs one 
and it might be reasonable approach to grasp the 
shape through the Gibbs form (effective temperature) of the distribution. 
%%%%%%%%%%%%%%%%%%%%%%%%%%%%%%%%%%%%%%%%%
\section{Formulation and tools}
%%%%%%%%%%%%%%%%%%%%%%%%%%%%%%%%%%%%%%%%%%
In this section, we explain our formulation 
and several tools 
to evaluate the average-case performance of GA 
through the effective temperature scheduling 
of the Gibbs distribution that is 
trained from gene configurations of simple GA. 
%%%%%%%%%%%%%%%%%%%%%%%%%%%%%%%%
\subsection{Kullback-Leibler information}
%%%%%%%%%%%%%%%%%%%%%%%%%%%%%% 
We start our argument from the distance 
between an empirical distribution 
from GA dynamics 
$P_{GA}^{(t)}(\bm{s})$
and a Gibbs distribution 
$P_{B}^{(t)}(\bm{s})$
at the effective temperature 
$T =\beta^{-1}$.  
The distance is measured by the following 
Kullback-Leibler information (KL) 
%%%%%%%%
\begin{equation}
KL(P_{GA} \| P_{B})  =  
\sum_{\bm{s}}
P_{GA}(\bm{s})
\log 
\left\{
\frac{P_{B}(\bm{s})}
{P_{SA}(\bm{s})}
\right\}
\end{equation}
%%%%
where the summation with respect to 
all possible gene configurations 
$\bm{s}=(s_{1},\cdots,s_{N})$
is defined by 
%%%%%
%%%%%%%%%%
\begin{equation}
\sum_{\bm{s}} (\cdots) \equiv 
\sum_{s_{1}=\pm 1}
\cdots\sum_{s_{N}=\pm 1}
(\cdots).
\end{equation}
%%%%%%%%%%
In this paper, we represent each component of 
gene configurations 
by $s_{i} = \pm 1$ instead of $s_{i}=0,1$ 
because we choose 
the cost function of spin glasses to be minimized 
as a benchmark test later on. 
The `spin' here means a tiny magnet in atomic scale-length and 
$s_{i}=+1$ stands for `up-spin' and 
vice versa.  
We should keep in mind that the above 
distance is dependent on the 
inverse temperature $\beta$. 
Thus, we obtain 
the following Boltzmann-machine-type 
learning equation 
with respect to $\beta$ as 
%%%%%%
\begin{eqnarray}
\frac{d \beta}{dt} & = &  
-\frac{\partial KL(P_{GA}^{(t)} \| P_{B}^{(t)})}
{\partial \beta} =  
\sum_{\bm{s}}
P_{GA}^{(t)}(\bm{s})
\cdot 
\frac{
{\partial P_{B}^{(t)}(\bm{s})}
/{\partial \beta}}
{P_{B}^{(t)}(\bm{s})}. \nonumber \\
\label{eq:LE}
\end{eqnarray}
%%%%%
We naturally expect that 
the effective temperature evolves so as to 
minimize the KL information 
for each time step. 
When both distributions 
become identical one 
in the limit of $t \to \infty$, 
namely, $P_{GA}^{(\infty)}
(\bm{s})=
P_{B}^{(\infty)}(\bm{s})$, 
we obtain 
%%%%%%
\begin{eqnarray}
\frac{d \beta}{dt} & = &   
\sum_{\bm{s}}
P_{GA}^{(\infty)}
(\bm{s}) \cdot 
{
\{{\partial P_{B}^{(\infty)}(\bm{s})}
/{\partial \beta}}\}
/{P_{B}^{(\infty)}(\bm{s})} \nonumber \\
\mbox{} & = &   
({\partial}/{\partial \beta})
\sum_{\bm{s}}
P_{B}^{(\infty)}
(\bm{s}) \nonumber \\
\mbox{} & = & 
({\partial}/{\partial \beta})
\sum_{\bm{s}}
\delta (\bm{s}-
\bm{s}_{*}) =  
{\partial \alpha}/{\partial \beta}=0
\end{eqnarray}
%%%%%%%
and the time evolution of inverse-temperature then stops. 
We should notice that $\alpha \equiv \sum_{\bm{s}}
\delta (\bm{s}-
\bm{s}_{*})$ is the number of 
degeneracy at the lowest energy states. 
%%%%%%%%%%%%%%%%%%%%%%%%%%%%%%%%%%%%%%%%
\subsubsection{The Holland's condition}
%%%%%%%%%%%%%%%%%%%%%%%%%%%%%%%%%%%%%
Before we examine the time-dependence of the effective temperature 
$\beta$, we comment on the 
validity of the choice of a Gibbs form 
as the distribution. 
John Holland mentioned that 
the algorithm might be effective if 
the probability 
$P(\mathcal{H},t)=\sum_{i \in \mathcal{H}}p_{i}(t)$ 
that a schema $\mathcal{H}$ appears 
at generation (time step) 
$t$ follows the following condition 
as a kind of `Master equation' 
of probabilistic flow:  
%%%%%
\begin{equation}
\frac{dP(\mathcal{H},t)}{dt} =  
f(\mathcal{H},t)-P(\mathcal{H},t)f(\mathcal{J},t)
\label{eq:holland1}
\end{equation}
%%%%%
where $\mathcal{J}$ stands for an arbitrary schema 
which is different from the $\mathcal{H}$ and 
$p_{i}(t)$ is a 
probability that 
a gene configuration $i$ appears 
at generation $t$ \cite{Holland}. 
%%%%%%%%%%%%%%%%%%%%%%%%%%%%%
$f(\mathcal{H},t)$ denotes the average 
fitness of the schema $\mathcal{H}$ at generation $t$: 
%%%%%%%
\begin{equation}
f(\mathcal{H},t) =  
\sum_{i \in \mathcal{H}} g(i) p_{i}(t). 
\label{eq:skima}
\end{equation}
%%%%
The above equation means 
that the probability that a $\mathcal{H}$ 
appears increases proportional to 
the average fitness value of $\mathcal{H}$ 
and it also decreases proportional to 
the average fitness values $f(\mathcal{J},t)
\equiv \sum_{i \in \mathcal{J} \neq \mathcal{H}}g(i)p_{i}(t)$. 

One can easily show that 
the above condition is satisfied by a 
Gibbs distribution having the form: 
%%%%%%
\begin{equation}
p_{i}(t)  =  
\frac{{\exp}[\beta_{t} g(i)]}
{\sum_{j \in  \mathcal{J}}{\exp}[\beta_{t}g(j)]}
\label{eq:holland2}.
\end{equation}
%%%%
For simplicity, we assume that 
the inverse temperature increases 
linearly in time as $\beta_{t}=t$. 
Then, the above (\ref{eq:holland2}) leads to 
%%%%%%%
\begin{equation}
p_{i}(t) = 
\frac{{\exp}[t g(i)]}
{\sum_{j \in  \mathcal{J}}
{\exp}[t g(j)]}. 
\end{equation}
%%%%
Taking the derivative of both sides of 
the above equation 
with respect to $t$, we have 
%%%%%%%
\begin{eqnarray}
\frac{dp_{i}(t)}{dt} & = & 
\frac{
g(i)\,{\rm e}^{t g(i)}(
\sum_{j \in \mathcal{J}}{\rm e}^{t g(j)})
-{\rm e}^{t g(i)}
\sum_{j \in \mathcal{J}}g(j){\rm e}^{t g(j)}}
{
(\sum_{j \in \mathcal{J}}{\rm e}^{t g(j)})^{2}} \nonumber \\
\mbox{} & = & 
\frac{g(i) {\rm e}^{t g(i)}}
{\sum_{j \in \mathcal{J}}
{\rm e}^{t g(j)}}
-
\left(
\frac{{\rm e}^{t g (i)}}
{\sum_{j \in \mathcal{J}}
{\rm e}^{t g(j)}}
\right) \nonumber \\
\mbox{} & \times & 
\left(
\frac{\sum_{j \in \mathcal{J}}g(j) {\rm e}^{t g (i)}}
{\sum_{j \in \mathcal{J}}
{\rm e}^{t g(j)}}
\right) \nonumber \\
\mbox{} & = &  
p_{i}(t)
g(i)-
p_{i}(t)\sum_{j \in \mathcal{J}}g(j)p_{j}(t). 
\label{eq:holland3}
\end{eqnarray}
%%%%%%
Taking the derivative of 
$P(\mathcal{H},t)=\sum_{i \in \mathcal{H}}p_{i}(t)$ 
with respect to $t$ and 
substituting 
the above (\ref{eq:holland3}) 
into the right hand side of the equation, 
we obtain 
%%%%%%%%%
\begin{eqnarray}
\frac{d P(\mathcal{H},t)}{dt} & = & 
\sum_{i \in \mathcal{H}}
\frac{d p_{i}(t)}{dt} \nonumber \\
\mbox{} & = &  
\sum_{i \in \mathcal{H}}
\left\{
p_{i}(t)
g(i)-
p_{i}(t)\sum_{j \in \mathcal{J}}g(j)p_{j}(t)
\right\} \nonumber \\
\mbox{} & = & 
\sum_{i \in \mathcal{H}}
p_{i}(t)g(i) - 
\sum_{i \in \mathcal{H}}
p_{i}(t) 
\sum_{j \in \mathcal{J}}g(j)p_{j}(t) \nonumber \\
\mbox{} & = &   f(\mathcal{H},t)
-P(\mathcal{H},t)f(\mathcal{J},t)
\end{eqnarray}
%%%%%%%%%%%%
where we used the definition (\ref{eq:skima}) of 
average fitness of the 
schema $\mathcal{H}$ at generation $t$. 
This equation is nothing but  
the Holland's condition (\ref{eq:holland1}). 
This result means that 
the empirical distribution of 
genes which are generated by GA dynamics 
is more likely to be a Gibbs distribution 
or can be well-approximated by a Gibbs distribution 
specified by the inverse temperature $\beta$ if 
the GA effectively finds the solution for a 
given optimization problem. 
This fact provides 
us a justification of the present approach 
to make a Gibbs distribution learns from 
the GA dynamics. 
%%%%%%%%%%%%%%%%%%%%%%%%%%%%%%%
\subsection{Learning equation for spin systems}
%%%%%%%%%%%%%%%%%%%%%%%%%%%%%%%
In the previous section, we formulated the learning equation 
for general problems and discussed some key properties 
including the Holland's condition in the formulation. 
Here we attempt to restrict ourselves to 
more particular problems, 
namely, we deal with a class of combinatorial 
optimization problems whose cost functions 
are described by the energy function of Ising model.
 
We first reformulate the equation (\ref{eq:LE}) by means 
of Ising spin systems having the energy function 
$H(\bm{s}) = 
-\sum_{ij}
J_{ij}s_{i}s_{j}$. 
For the case of positive 
constant spin-spin interaction 
$J_{ij}=J>0,\,\,\,\forall_{i,j}$, 
the lowest energy state is 
apparently given by $s_{i}=+1,\,\,\,\forall_{i}$ (all-up spins) 
or $s_{i}=-1,\,\,\,\forall_{i}$ (all-down spins). 
However, as we shall see in the 
following sections, for 
the case of 
randomly distributed 
$J_{ij}$ (the $\pm$ sign is also random), 
the lowest energy state is 
highly degenerated and it becomes very hard to find the state.
It should be noted that 
the traveling salesman problem (TSP) (see e.g. \cite{Mezard}) 
or the $k$-satisfiability problem ($k$-SAT) (see e.g. \cite{Monasson}) 
is rewritten 
in terms of optimization problems 
described by the variant of 
the above energy function of spin glasses. 

Substituting the corresponding Gibbs distribution 
$P_{B}(\bm{s})=
{\exp}[-\beta H(\bm{s})]/
\sum_{\bm{s}}
{\exp}[-\beta H(\bm{s})]$ 
into equation (\ref{eq:LE}), the learning equation leads to 
%%%%%%%%%%%%%%
%%%%%
\begin{eqnarray}
&& \hspace{-2cm}\frac{d\beta}{dt}  = 
\sum_{\bm{s}}
P_{GA}(\bm{s})
\left(
\sum_{ij}J_{ij}
s_{i}s_{j}
\right) \nonumber \\
\mbox{} & - &   
\frac{\sum_{\bm{s}}
(\sum_{ij}J_{ij}s_{i}s_{j})\,
{\exp}[\beta \sum_{ij}J_{ij}s_{i}s_{j}]
}
{\sum_{\bm{s}}
{\exp}[\beta \sum_{ij}
J_{ij}s_{i}s_{j}]
}
\label{eq:learning}
\end{eqnarray}
%%%%%%%%%%%%%%%
where the second term appearing in 
the right hand side of 
the above equation is internal energy 
of the system described by the Hamiltonian 
$H(\bm{s}) = 
-\sum_{ij}
J_{ij}s_{i}s_{j}$ at 
temperature $T=\beta^{-1}$, 
whereas the first term is the energy 
$H(\bm{s})$ 
averaged over 
the empirical distribution $P_{GA}(\bm{s})$ of GA. 
Then, we immediately find that the condition 
%%%%
%%%%%%
\begin{eqnarray}
\mbox{}&& \hspace{-1.5cm} \sum_{\bm{s}}
P_{GA}(\bm{s})
(
\sum_{ij}J_{ij}
s_{i}s_{j})  =   
\sum_{\bm{s}}
P_{B}(\bm{s})
(
\sum_{ij}J_{ij}
s_{i}s_{j}) \nonumber \\
\mbox{} & = & 
\frac{\sum_{\bm{s}}
(\sum_{ij}J_{ij}s_{i}s_{j})
\,{\exp}[\beta \sum_{ij}J_{ij}s_{i}s_{j}]
}
{\sum_{\bm{s}}
{\exp}[\beta \sum_{ij}
J_{ij}s_{i}s_{j}]
}
\end{eqnarray}
%%%%%%%%%%%%
yields $d\beta/dt =0$
for 
$P_{GA}(\bm{s})
=P_{B}(\bm{s})$. 

In general, it is very hard to 
calculate the internal energy of the 
spin system 
%%%%
\begin{equation}
U(\{J\}:\beta) \equiv  
-\frac{\sum_{\bm{s}}
(\sum_{ij}J_{ij}s_{i}s_{j})\,
{\exp}[\beta \sum_{ij}J_{ij}s_{i}s_{j}]
}
{\sum_{\bm{s}}
{\exp}[\beta \sum_{ij}
J_{ij}s_{i}s_{j}]
}
\label{eq:EJbeta}
\end{equation}
%%%%%
because $2^{N}$ sums 
for all possible configurations in 
$\sum_{\bm{s}}(\cdots)$ 
are needed to evaluate the $E(\{J\} :\beta)$, 
where we defined a set of interactions by 
%%%%
\begin{equation}
\{J\} \equiv 
\{
J_{ij}|i,j=1,\cdots,N
\}. 
\end{equation} 
To overcome this difficulty, we usually 
use the so-called Markov chain Monte Carlo 
(MCMC) method to calculate the expectation 
(\ref{eq:EJbeta}) by important sampling from 
the Gibbs distribution 
at temperature $T =\beta^{-1}$. 

On the other hand, 
the first term appearing in the right hand 
side of (\ref{eq:learning}), 
we evaluate the expectation by making use of 
%%%%%%
\begin{eqnarray}
&& \hspace{-1.5cm} U_{GA}(\{J\}) \equiv  
-\sum_{\bm{s}}
P_{GA}(\bm{s})
\left(
\sum_{ij}J_{ij}
s_{i}s_{j}  
\right) \nonumber \\
\mbox{} & = & 
-\lim_{L \to \infty}
\frac{1}{L}
\sum_{l=1}^{L}
\left(
\sum_{ij}
J_{ij}
s_{i}(t,l)
s_{j}(t,l)
\right)
\end{eqnarray}
%%%%
where $s_{i}(t,l)$ is 
the $l$-th sampling point at time $t$ from 
the empirical distribution of 
GA. Namely, we shall replace the expectation of 
the cost function 
$H(\bm{s})=-\sum_{ij}J_{ij}s_{i}s_{j}$ 
over the distribution $P_{GA}(\bm{s})$ 
by sampling from the empirical distribution of GA. 

By a simple transformation $\beta  \to T^{-1}$ 
in equation (\ref{eq:learning}), 
we obtain the Boltzmann-machine-type 
learning equation with respect to 
effective temperature $T$ as follows. 
%%%%%%
\begin{eqnarray}
\frac{dT}{dt} & = & 
-T^{2}
\left(
U(\{J\}: T^{-1})
-U_{GA}(\{J\})
\right)
\label{eq:learning2}
\end{eqnarray}
%%%%%%%%
From this learning equation, 
we find that time-evolution of 
effective temperature depends on 
the difference 
between 
the expectations of 
the cost function over 
the Gibbs distribution 
at temperature $T$ and 
the empirical distribution of GA.  

Obviously, the performance 
of GA is now evaluated through 
the `annealing schedule' of 
effective temperature $T$, however, 
the schedule depends on 
the choice of interactions 
between spins, that is, $\{J\}$. 
Therefore, 
we should average the learning 
equation (\ref{eq:learning2}) over the 
such problem-dependent `input data', that is to say, $\{J\}$.
%%%%%%%%%%%%%%%%%%%%%%%%%%%%%%%%%%
\subsection{Average-case performance}
%%%%%%%%%%%%%%%%%%%%%%%%%%%%%%%%%%%
As we mentioned, the difficulties of 
finding the lowest energy states 
depend on the weights 
between spins, namely, 
the problem is dependent on 
the statistical properties of interactions $\{J \}$. 
 Obviously, 
 the learning equation  
 (\ref{eq:learning2}) and 
 its time evolution for 
 a finite size system 
 depends on the choice of 
 $\{J \}$.  
 Hence, 
 the GA which is applied to some specific problem 
 having a set of $\{J \}$ 
 might  give an excellent solution 
 as a peculiar case and 
 the reverse might be also true (the GA might give a poor solution 
 as another peculiar case). 
Therefore, 
we should evaluate the `average-case 
performance' of the learning 
equation 
which is independent of 
the realization of `problem' $\{J \}$. 
Namely, one should evaluate the `data-averaged' learning equation 
%%%%%
%%%%%%
\begin{equation}
\frac{dT}{dt} =  
-T^{2}
\left(
\mathbb{E}_{\{J\}}
\left(
U(\{J\}: T^{-1})
\right)  - 
\mathbb{E}_{\{J\}}
\left(
U_{GA}(\{J\})
\right) 
\right)
\label{eq:learning3}
\end{equation}
%%%%%%%%
to discuss the 
average-case performance, 
where we defined the average 
$\mathbb{E}_{\{J\}}(\cdots)$ by 
%%%%%%%%
\begin{equation}
\mathbb{E}_{\{J\}} (\cdots)  \equiv 
\prod_{ij} 
\int 
dJ_{ij}
(\cdots)
P(J_{ij})
\label{eq:def_ave}. 
\end{equation}
%%%%
We should keep in mind that 
in this paper we deal with the problem 
in which 
each interaction $J_{ij}$ 
has no correlation with the others, namely, 
%%%%
\begin{equation}
\mathbb{E}_{\{J\}}(J_{ij}J_{kl}) = 
J^{2} \delta_{i,k}\delta_{j,l}
\end{equation}
%%%
where we defined 
$J^{2}$ as a variance of $P(J_{ij})$ and 
$\delta_{x,y}$ stands for 
a Kronecker's delta. 
%%%%%%%%%%%%%%%%%%%%%%%%%%%%%%%%
\subsubsection{Self-averaging of physical quantities}
%%%%%%%%%%%%%%%%%%%%%%%%%%%%%%%%
In order to carry out the performance 
evaluation, 
we need to calculate the average of 
equation (\ref{eq:learning2}) over the probability of 
realization $\{J \}$, that is, 
%%%%%%%%%%%%%%%%%%
%%%%%%
\begin{eqnarray}
\frac{dT}{dt} & = & T^{2}\lim_{L \to \infty}
\frac{1}{L}
\sum_{l=1}^{L}
\mathbb{E}_{\{J\}}
\left(
\sum_{ij}
J_{ij}
s_{i}(t,l)
s_{j}(t,l)
\right) \nonumber \\
\mbox{} & - &  
T^{2}
\mathbb{E}_{\{J\}}
(U(\{J\}:\beta)). 
\label{eq:result02}
\end{eqnarray}
%%%%%%%%
In statistical physics of disordered spin systems, 
the probability that 
an arbitrary state $x$ 
having the energy $U_{x}$ 
appears is given 
by $P_{x}={\exp}[-\beta U_{x}]/Z$ 
where a normalization 
factor $Z$ is referred to as 
{\it partition function} 
%%%%
\begin{equation}
Z = 
\sum_{x}{\exp}[-\beta U_{x}].
\end{equation}
%%%%  
Then, the internal energy defined by 
%%%%%
\begin{equation}
U = 
\frac{\sum_{x}U_{x}\,{\exp}[-\beta U_{x}]}
{\sum_{x}{\exp}[-\beta U_{x}]} = 
\frac{\sum_{U_{x}}\mathcal{D}(U_{x})
U_{x}\,{\exp}[-\beta U_{x}]}
{\sum_{U_{x}}\mathcal{D}(U_{x})
{\exp}[-\beta U_{x}]}, 
\end{equation}
%%%%
where $\mathcal{D}(U_{x})$ stands for a density 
of state having the energy $U_{x}$, is 
obtained from 
the free energy 
%%%%%
\begin{equation}
F = -T \log Z
\end{equation}
by using the following relation 
%%%%
\begin{equation}
U =\frac{\partial}{\partial \beta}
(\beta F).
\end{equation}
%%%%%%%%%%%%
To use the relationship  
between the internal and free energies, 
one can rewritten the second term 
appearing in the right hand side of 
the above equation (\ref{eq:result02}) as 
%%%%
\begin{equation}
\mathbb{E}_{\{J\}}(
U(\{J\}:\beta)) = 
\frac{\partial}{\partial \beta}
\left(
\beta 
\mathbb{E}_{\{J\}}
(F(\{J\} : \beta))
\right).
\end{equation}
%%%%
In statistical physics 
of disordered spin systems, 
it is well-known that 
the quantities such as free energy 
are independent of the choice of $\{J \}$ 
in the large system size limit $N \to \infty$. 
In other words, 
the free energy calculated 
for a given realization of 
$\{J \}$ 
is identical to 
the average over the 
probability 
$P(\{J \})=\prod_{ij}P(J_{ij})$, 
namely, the identity 
%%%%%
\begin{equation}
\lim_{N \to \infty}
F(\{J\}_{\mbox{\tiny a realization}} : \beta) =  
\mathbb{E}_{\{J\}} \left(
F(\{J\} : \beta)
\right)
\label{eq:self}
\end{equation}
%%%%
holds. 
The mathematically rigorous 
proof for the Sherrington-Kirkpatrick model 
is given elsewhere (see e.g. \cite{Talagrand}). 
Thus, we calculate 
the right hand side of 
(\ref{eq:self}) for 
mathematically solvable model, whereas we 
evaluate the left hand side by computer simulations 
for the other models. 
However, as we mentioned above, the both procedures to 
evaluate the average give the same 
results in the limit of $N \to \infty$. 
%%%%%%%%%%%%%%%%%%%%%%%%%%%%%%%%%%%%%%
\subsubsection{The replica method}
%%%%%%%%%%%%%%%%%%%%%%%%%%%%%%%%%%%%%%
Here we encounter a technical problem in the evaluation of the average. 
As we mentioned, we should evaluate the average 
such as $\mathbb{E}_{\{J\}} (\cdots)$, 
namely, the quantity to be evaluated is now written as follows.  
%%%%%%%
\begin{eqnarray}
&& \hspace{-1.5cm} U  =  
\mathbb{E}_{\{J\}} \left(
\frac{\partial}{\partial \beta}
(\beta F)
\right) = \frac{\partial}{\partial \beta}
(\beta \mathbb{E}_{\{J\}}(F)) \nonumber \\
\mbox{} & = & 
\frac{\partial}{\partial \beta}
\mathbb{E}_{\{J\}} 
\log  \biggl( 
\sum_{\bm{s}}
{\exp}[-\beta H(\bm{s} : \{J\})] \biggr)
\end{eqnarray}
%%%%
Unfortunately, 
it is very difficult for us to carry out the above 
calculation except for 
a few limited cases 
because the 
variables 
$\{J\}$ appear in the logarithm of 
the partition function. 
Then, 
by making use of the 
identity: $\log Z = (Z^{n}-1)/n$ which 
holds in the limit of $n \to 0$, 
we calculate the average as  
%%%%%
\begin{eqnarray}
&& \hspace{-1.5cm} \mathbb{E}_{\{J\}} (\log Z)  = 
\lim_{n \to 0}
\frac{\mathbb{E}_{\{J\}}(Z^{n})-1}{n} \nonumber \\
\mbox{} & = &  
\lim_{n \to 0}
\frac{
\mathbb{E}_{\{J\}}(
\prod_{a=1}^{n}
\sum_{\bm{s}_{a}}
{\rm e}^{-\beta \sum_{a}H(\bm{s}_{a} : \{J\})} 
)-1}{n}
\end{eqnarray}
%%%%%%
where we replaced the average of 
$\log Z$, namely, 
$\mathbb{E}_{\{J\}}(\log Z)$ 
with the average of $Z^{n}$, that is 
$\mathbb{E}_{\{J\}}(Z^{n})$
by introducing 
the $n$-replicas (copies) $a=1,2,\cdots, n$. 
This procedure to calculate the average 
of self-averaging quantities 
is referred to as {\it replica method} \cite{SK,SpinGlass}. 
In the evaluation of 
the learning equation 
for the problem having the cost function of 
the Sherrington-Kirkpatrick-type, 
we shall use this technique.  
%%%%%%%%%%%%%%%%%%%%%%%%%%%%%%%%%%%%%%%
\section{Mathematically tractable models}
%%%%%%%%%%%%%%%%%%%%%%%%%%%%%%%%%%%%%%%%
In this section, we introduce 
two kinds of spin glass model 
which will be used as a benchmark cost function 
to be minimized by GA. 
These models are very simple, 
however, several quantities 
such as internal energy as a function of 
temperature are 
obtained analytically and 
very suitable for us to examine 
the average-case performance of 
GA as a benchmark test.  
The models dealt with are given as follows.  
%%%%%%%%%%%%%%%%%%%%%%%%%%%%%%%%%%%%%%%%%%%
\begin{itemize}
\item
Spin glass chain\\
It is one-dimensional spin glass model 
having only nearest neighboring interactions．
It is possible for us to 
investigate the temperature dependence of 
internal energy and 
moreover, 
one can obtain the lowest energy 
exactly. 
The energy function (Hamiltonian 
in the literature of statistical physics) is given by 
%%%%%
\begin{eqnarray}
H & = & -\sum_{i=1}^{N}
J_{i}s_{i}s_{i+1},\,\,\,\,
J_{i}={\cal N}(0,1)
\label{eq:1dimSK}
\end{eqnarray}
%%%%%
where 
$J_{i}$ stands for the interaction between 
spins $s_{i}$ and $s_{i+1}$.  $\mathcal{N}(a,b)$ 
denotes 
a normal Gaussian distribution 
with mean $a$ variance $b$．
%%%%%%%%%%%%%%%%%%%%%%%%%%
%%%%%%%%%%%%%%%%%%%%%%%%%%%
\begin{figure}[ht]
\begin{center}
\includegraphics[width=8cm]{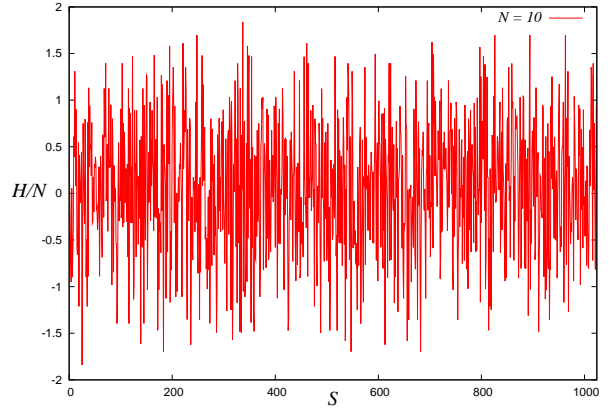}
\end{center}
\caption{\footnotesize 
Typical energy landscape 
$H(\bm{s})=-\sum_{i}J_{i}s_{i}s_{i+1}$ 
with $P(J_{i})=\mathcal{N}(0,1)$, 
$\mathbb{E}(J_{i}J_{j})=\delta_{i,j}$   
of 
the spin glass chain. 
The number of spins is $N=10$.  
It should be noted that 
the horizontal axis $S$ denotes the label of 
states, 
that is, 
$S=1,2,\cdots, 2^{N} (=1028)$. 
For instance, 
$S=1$ stands for a state, 
say, $\bm{s}(S=1)=
(+1,+1,\cdots,+1)$ and 
$S=2^{N}$ denotes 
$\bm{s}(S=2^{N})=
(-1,-1,\cdots,-1)$.  
}
\label{fig:energy_dist_SGC}
\end{figure}
%%%%%%%%%%%%%%%%%%%%
We plot the typical energy landscape 
in Figure \ref{fig:energy_dist_SGC}. 
From this figure, we find that 
the structure of the energy surface is 
complicated and it seems to be difficult for us 
to find the lowest energy state. 

However, we should notice that 
in (\ref{eq:1dimSK})
$s_{i}$ takes 
$\pm 1$ and the product 
$s_{i}s_{i+1}$ also has a value 
$\pm 1$. 
Hence, we introduce the new variable 
$\tau_{i}$ which is defined by 
$\tau_{i}=s_{i}s_{i+1}$, 
then $\tau_{i}$ takes 
$\tau_{i} \in \{1,-1\}$. 
Therefore, 
in order to minimize  
$H (\bm{\tau}) =  -\sum_{i}J_{i}\tau_{i}$, 
we should 
determine 
$\tau_{i}={\rm sgn}(J_{i})$
for each $i$ and 
then, we have 
the lowest energy as 
$U_{\rm min} = 
-\sum_{i}
J_{i} \, {\rm sgn}(J_{i}) = 
-\sum_{i}|J_{i}|$. 
%%%%%%%%%%%%%%%%%%%%%%%%
\begin{figure}[ht]
\begin{center}
\includegraphics[width=8cm]{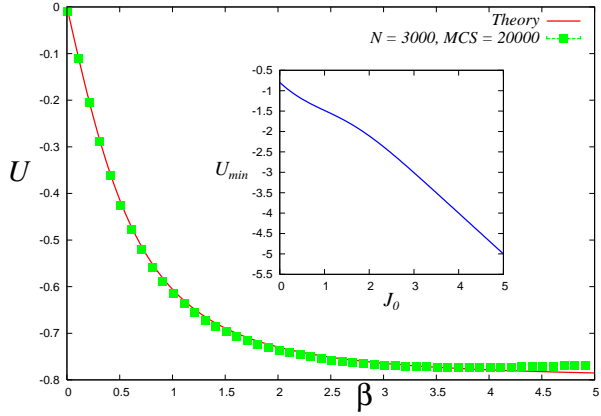}
\end{center}
\caption{\footnotesize
Internal energy of spin glass chain 
as a function of 
temperature. 
The solid line is 
exact result 
$U = -\beta 
\int_{-\infty}^{\infty}
\frac{Dx}
{\cosh^{2}\beta x}$, 
whereas the dots 
denote 
the internal energy calculated by 
the MCMC for 
$N=3000$. 
The error-bars are calculated by 
$10$-independent runs for 
different choice of 
the 
$\{J\} \equiv \{J_{i}|i=1,\cdots,N\}$. 
The inset indicates 
the 
$U_{\rm min}$ as a function of $J_{0}$. 
We set $J=1$. 
}
\label{fig:fg1}
\end{figure}
%%%%%%%%%%%%%%%%%%%%%
Namely, 
when   
$J_{i}$ obeys 
a Gaussian with 
mean 
$J_{0}$ and 
variance $J^{2}$, 
the lowest energy 
for a single spin is obtained in the 
thermodynamic limit  
$N\to \infty$ as 
%%%%%%
\begin{eqnarray*}
\lim_{N \to \infty}
\frac{U_{\rm min}}{N} & = & 
\mathbb{E}_{\{J\}}(
|J_{i}|)=
\int_{-\infty}^{\infty}
\frac{dJ_{i}}{\sqrt{2\pi}J}
\,
{\rm e}^{-\frac{(J_{i}-J_{0})^{2}}{2J^2}}
|J_{i}| \nonumber \\
\mbox{} & = & 
-J_{0}-J
\sqrt{\frac{2}{\pi}}
\,{\rm e}^{-\frac{J_{0}^2}
{2J^{2}}}
\end{eqnarray*}
%%%%%
where  
$\mathbb{E}_{\{J\}}(\cdots)$ 
here stands for the average over the 
configuration
$\{J\}\equiv(J_{1},\cdots,J_{N})$. 

Thus, 
for the choice of 
$(J_{0},J)=(1,0)$, 
namely, 
in the limit of 
the ferromagnetic Ising model, 
we have the lowest energy as 
$U_{\rm min}/N=-1$ 
(all spins align in the same direction), 
On the other hand, for the choice of 
$(J_{0},J)=(0,1)$, 
we have 
$U_{\rm min}=-\sqrt{2/\pi}$. 
These facts mean that 
the lowest energy changes according to 
the value of  ratio $J_{0}/J$.

We next consider the case of 
finite temperature ($\beta < \infty$). 
For this case 
internal energy per spin 
is given by 
%%%%%%
\begin{equation}
\lim_{N \to \infty}
\frac{\langle H \rangle_{\tau}}
{N}  = \mathbb{E}_{\{J\}}(\langle H \rangle_{\tau}) = 
-\frac{\partial}{\partial \beta}
\log 
\sum_{\bm{\tau}}
{\rm e}^{\beta \sum_{i}J_{i}\tau_{i}}
\end{equation}
%%%
with 
%%%%%
\begin{equation}
\langle \cdots \rangle_{\tau} \equiv 
\frac{\sum_{\bm{\tau}}
{\exp}[\beta \sum_{i}J_{i}\tau_{i}]}
{Z_{\tau}}
\end{equation}
%%%%%
where 
we defined 
%%%%
\begin{equation}
\sum_{\bm{\tau}}(\cdots) \equiv 
\sum_{\tau_{i}=\pm 1}
\cdots \sum_{\tau_{N}=\pm 1}(\cdots)
\end{equation}
%%%% 
and the partition function $Z_{\tau}=\sum_{\bm{\tau}}
{\rm e}^{\beta \sum_{i}J_{i}\tau_{i}}$ 
is now calculated as 
$\{
2\cosh (\beta J_{i})\}^{N}$. 
Hence, we have the average free energy 
density 
defined by  
$f=\lim_{N \to \infty}
(\log Z/N)=N^{-1}\mathbb{E}_{\{J\}} (\log Z)$ 
is evaluated as follows (the self-averaging property 
we mentioned before was assumed). 
%%%%%
\begin{eqnarray}
f & = & 
\int_{-\infty}^{\infty}
\frac{dJ_{i}}
{\sqrt{2\pi}\,J}
{\rm e}^{-\frac{(J_{i}-J_{0})^{2}}{2J^{2}}}
\log 2 \cosh (\beta J_{i}) \nonumber \\
\mbox{} & = & 
\int_{-\infty}^{\infty}
Dx \log 2 \cosh \beta (J_{0}+Jx) 
\end{eqnarray}
%%%%%%
where we defined 
$Dx \equiv dx\,{\rm e}^{-x^{2}/2}/\sqrt{2\pi}$.  
From the above result, we immediately obtain 
the internal energy per spin 
$U=-{\partial f}/{\partial \beta}$ by 
%%%%%%
\begin{eqnarray}
U & = & 
-J_{0}
\int_{-\infty}^{\infty}
Dx \tanh 
\beta (J_{0}+Jx) \nonumber \\
\mbox{} & - & 
\beta J^{2} 
\int_{-\infty}^{\infty}
\frac{Dx}
{\cosh^{2}\beta (J_{0}+Jx)}. 
\end{eqnarray}
%%%%%
Especially, for 
the case of 
$(J_{0},J)=(0,1)$,  we have 
%%%%%%
\begin{equation}
U = 
-\beta 
\int_{-\infty}^{\infty}
\frac{Dx}
{\cosh^{2}\beta x}.
\end{equation}
%%%%%%%%%%%%% 
In Figure \ref{fig:fg1}, we show the 
$U$ as a function of $T$. 
%%%%%
From the arguments 
we provided above, 
we have the following 
learning 
equation (\ref{eq:result2}) 
for the spin glass chain 
whose Hamiltonian 
is given by (\ref{eq:1dimSK}) is now rewritten as 
%%%%%%
\begin{eqnarray}
\frac{dT}{dt} & = & 
T^{2}\lim_{L \to \infty}
\frac{1}{L}
\sum_{l=1}^{L}
\left(
\sum_{i}
J_{i}
s_{i}(t,l)
s_{i+1}(t,l)
\right) \nonumber \\
\mbox{} & - & 
T
\int_{-\infty}^{\infty}
\frac{Dx}
{\cosh^{2}T^{-1}x}.
\label{eq:result2}
\end{eqnarray}
%%%%%
%%%%%%%%%%%%%%%%%%%%%%%%%%
%%%%%%%%%%%%%%%%%%%%%%%%%%
\item
Sherrington-Kirkpatrick model\\
%%%
This model is 
a spin glass model in which 
each spin is located on a 
complete graph. For this model, 
the energy function is explicitly 
given by 
%%%%%%%%%%%%
\begin{equation}
H = 
-\frac{1}{N}\sum_{i=1}^{N}
\sum_{j\neq i}
J_{ij}s_{i}s_{j}
\label{eq:SK}
\end{equation}
%%%%%
where 
$J_{ij}$
obeys  $P(J_{ij}) = 
{\cal N}(J_{0},J^{2})$. 
%%%%%%%%%%%%%%%%%%%%%%%%%%%
\begin{figure}[ht]
\begin{center}
\includegraphics[width=8cm]{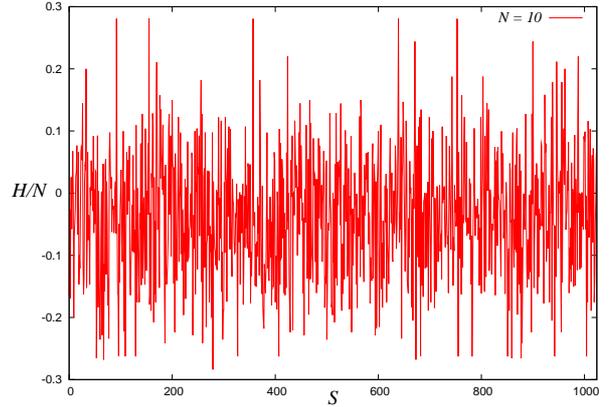}
\end{center}
\caption{\footnotesize 
Typical energy landscape 
$H(\bm{s})=-\sum_{ij}J_{ij}s_{i}s_{j}$ 
with $P(J_{ij})=\mathcal{N}(0,1)$, 
$\mathbb{E}(J_{ij}J_{kl})=\delta_{i,k}\delta_{j,l}$ 
of the SK model. 
The number of spins is $N=10$. 
It should be noted that 
the horizontal axis $S$ denotes the label of 
states, 
that is, 
$S=1,2,\cdots, 2^{N} (=1028)$. 
For instance, 
$S=1$ stands for a state, 
say, $\bm{s}(S=1)=
(+1,+1,\cdots,+1)$ and 
$S=2^{N}$ denotes 
$\bm{s}(S=2^{N})=
(-1,-1,\cdots,-1)$.  
 }
  \label{fig:energy_dist_SK}
\end{figure}
%%%%%%%%%%%%%%%%%%%%
We plot the typical energy landscape in 
Figure \ref{fig:energy_dist_SK}. 
At first glance, it seems that 
the structure of energy surface is 
very similar to that of the 
spin glass chain, 
however, finding the lowest energy state 
of the SK model needs much more difficult tasks. 
This `ground state problem' in the SK model 
is one of the non-trivial issues in the research field of spin glasses. 

By using the replica method 
we mentioned in the previous section, 
the averaged internal energy per spin, 
namely, 
the second term appearing 
in the 
learning equation 
(\ref{eq:result2}) 
is calculated as 
%%%%%
\begin{equation}
\frac{U}{N}  \equiv   U_{\beta}(m,q) =   
-\frac{J_{0}}{2}m^{2}-
\frac{\beta J^{2}}{2}
(1-q^{2})
\label{eq:SKU}
\end{equation}
%%%%%
where, $m, q$ are 
the replica symmetric 
solution 
for 
the magnetization and 
the spin glass order parameter, 
respectively. These are explicitly given by 
the following equations of state
%%%%%%%%%%%%%
\begin{equation}
m \equiv 
\frac{1}{N}
\sum_{i} \mathbb{E}_{\{J\}} (\langle s_{i}  \rangle) = 
\int_{-\infty}^{\infty} Dz\,\tanh 
\beta (Jz\sqrt{q}+J_{0}m) 
\label{eq:SKm} 
\end{equation}
\begin{equation}
q  \equiv 
\frac{1}{N}\sum_{i}
\mathbb{E}_{\{J\}}
(\langle s_{i} \rangle^{2})= 
\int_{-\infty}^{\infty} Dz\,\tanh^{2} 
\beta (Jz \sqrt{q}+J_{0}m)
\label{eq:SKq}
\end{equation}
%%%%%
where we defined 
%%%%
\begin{equation}
\langle \cdots 
\rangle \equiv 
\frac{\sum_{\bm{s}}
(\cdots) 
{\exp}[(\beta/N)\sum_{ij}s_{i}s_{j}]}
{\sum_{\bm{s}}
{\exp}[(\beta/N)\sum_{ij}s_{i}s_{j}]}
\end{equation}
%%%%%%
and $\mathbb{E}_{\{J\}} (\cdots)$ by (\ref{eq:def_ave}).  
For these solutions for 
a given temperature $T$, 
the learning equation for the SK model 
is obtained by 
%%%%%
%%%%%%
\begin{eqnarray}
\frac{dT}{dt} & = & 
T^{2}\lim_{L \to \infty}
\frac{1}{NL}
\sum_{l=1}^{L}
\left(
\sum_{i}
\frac{J_{ij}}{N}
s_{i}(t,l)
s_{j}(t,l)
\right) \nonumber \\
\mbox{} & - & 
T^{2}U_{\beta}(m(T),q(T)).
\label{eq:result_SK}
\end{eqnarray}
%%%%%
Here we should keep in 
mind that in the limit of 
$\beta \to \infty$,  
$\tanh \beta (\cdots)={\rm sgn}(\cdots)$, 
$q=1$ is derived from 
(\ref{eq:SKq}). 

On the other hand, 
from (\ref{eq:SKm}), 
the magnetization $m$ leads to 
%%%%%%%%%%%%%%%%%
\begin{equation}
m = 
\int_{-\infty}^{\infty}
Dz\,
{\rm sgn}
(Jz + J_{0}m) =   
1-2\,{\rm erfcc}
\left(
\frac{J_{0}}{J}m
\right)
\label{eq:SKm2}
\end{equation}
%%%%
where we defined ${\rm erfcc} (x)$ by 
%%%%%
\begin{equation}
{\rm erfcc} (x) \equiv  
\int_{x}^{\infty}
\frac{dz}{\sqrt{2\pi}}\,
{\rm e}^{-\frac{z^{2}}{2}}. 
\end{equation}
%%%%
By utilizing the asymptotic form 
$2 \,{\rm erfcc}(x) \simeq 1-x\sqrt{2/\pi}$ 
around $x \simeq 0$, 
we obtain the critical point 
$a \equiv \sqrt{2/\pi}\,(J_{0}/J)=1$ below which 
the spin glass phase emerges. 
Hence, the lowest energy 
at zero temperature 
is obtained by substituting the solution $m$ of 
(\ref{eq:SKm2}) into 
the expression of 
internal energy  
(\ref{eq:SKU}) with $q=1$, 
namely,  
$U/N =   - J_{0}m^{2}/2$. 
As a special case, 
the lowest energy in the ferromagnetic state 
$J_{0}/J  \to \infty$ is given by $U/J_{0}N  =   -1/2$. 
%%%%% 
In Figure \ref{fig:fgSKT0}, 
we plot the 
internal energies per spin 
scaled as $\sqrt{2/\pi}\,(U/JN)=-am^{2}/2$ and 
$U/J_{0}N=-m^{2}/2$ 
and 
magnetization 
$m$ as a function of $a$ ($\equiv \sqrt{2/\pi}\,(J_{0}/J)$) 
at $T=0$. 
%%%%%%%%%
\begin{figure}[ht]
\begin{center}
\includegraphics[width=8cm]{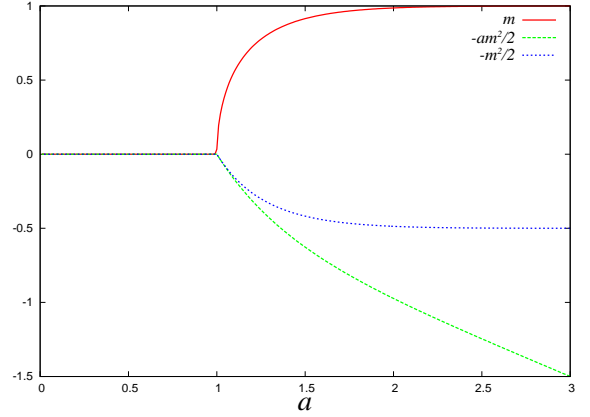}
\end{center}
\caption{\footnotesize 
Parameter $a$ ($\equiv \sqrt{2/\pi}\,(J_{0}/J)$)-dependence of 
magnetization $m$,  
internal energies per spin 
scaled as $\sqrt{2/\pi}(U/JN)=-am^{2}/2$ and 
$U/J_{0}N=-m^{2}/2$.}
\label{fig:fgSKT0}
\end{figure}
\end{itemize}
%%%%%
For these two mathematically tractable 
models, we shall evaluate the learning equations 
for effective Gibbs distributions  
in the next section. 
%%%%%%%%%%%%%%%%%%%%%%%%%%%%%%%%%%%%%%%%%%%%%%%%%%%%%
\section{Set-up for numerical experiments}
%%%%%%%%%%%%%%%%%%%%%%%%%%%%%%%%%%%%%%%%%%%%%%%%%%%
For these two kinds of 
the solvable spin glass models, 
we examine the 
learning equations (\ref{eq:result2})(\ref{eq:result_SK}) 
through the time-dependence of the effective temperature $T$．
In following, we explain our setting of 
parameters which 
control simple GA to be 
utilized in the learning processes. 
%%%%%%%%%%%%
\begin{itemize}
\item
{\bf The number of spins $N$}\\
This is the number of components 
in a single gene configuration and is 
regarded as the number of spins 
in the spin glass model. 
Here we set 
$N=2000$ for spin glass chain and 
$N=500$ for the SK model. 
%%%%
\item
{\bf The number of ensembles (population) $M$}\\
The number of population in GA. 
We set $M=100$ 
%%%%
\item
{\bf Parameters appearing in GA}
\begin{itemize}
\item
$\sigma$: The number of members  
in selection of tournament -type at each generation. 
\item
$p_{c}$:  The rate for a single point crossover 
\item
$p_{m}$:  The mutation rate 
\end{itemize}
%%%%
\item
{\bf Effective temperature $T$} \\
A control parameter of the Gibbs distribution 
to approximate the empirical distribution of GA. 
We set the initial value $T=T_{0}\,(<\infty)$．
%%%%
\item
{\bf On the selection} \\
In our numerical experiments, 
we generate the configurations (members) 
with length $N$ randomly and 
for each of the member, we evaluate the 
fitness values. 
Then, we pick up $\sigma$ 
members among the population (ensemble) 
with size $M$ and 
select the largest fitness member and 
the others are discarded. 
We repeat the process up to $M$ times. 

As another candidate of 
selection, we might use the method to 
weight each member $\alpha$ of 
population with 
$p_{\alpha}={\rm e}^{-\beta_{s}E_{\alpha}}/\sum_{\alpha=1}^{M}
{\rm e}^{-\beta_{s}E_{\alpha}}$ 
(see e.g. \cite{Bennett1,Bennett2}). 
Obviously, $\beta_{s} \to \infty$ 
limit yields the case in which only the 
best solution is selected. 
Hence, the case $\sigma=M$ 
for our selection rule 
is identical to the $\beta_{s} \to \infty$ limit. 
On the other hand, 
$\beta_{s}=0$ limit corresponds to 
$\sigma=1$, 
namely, each member is selected randomly. 
\end{itemize}
%%%%%%%%%%%%%%%%%%%%%%%%%%%
\section{Results of numerical experiments}
%%%%%%%%%%%%%%%%%%%%%%%%%%%%%%
According to 
the set-up explained in the 
previous section, we shall carry out the numerical experiments 
for two mathematically tractable models. 
The results are summed up below. 
%%%%%%%%%%%%%%%%%%%%%%%%%%%%%%%
\subsection{Spin glass chain}
\begin{figure}[ht]
\begin{center}
\includegraphics[width=8cm]{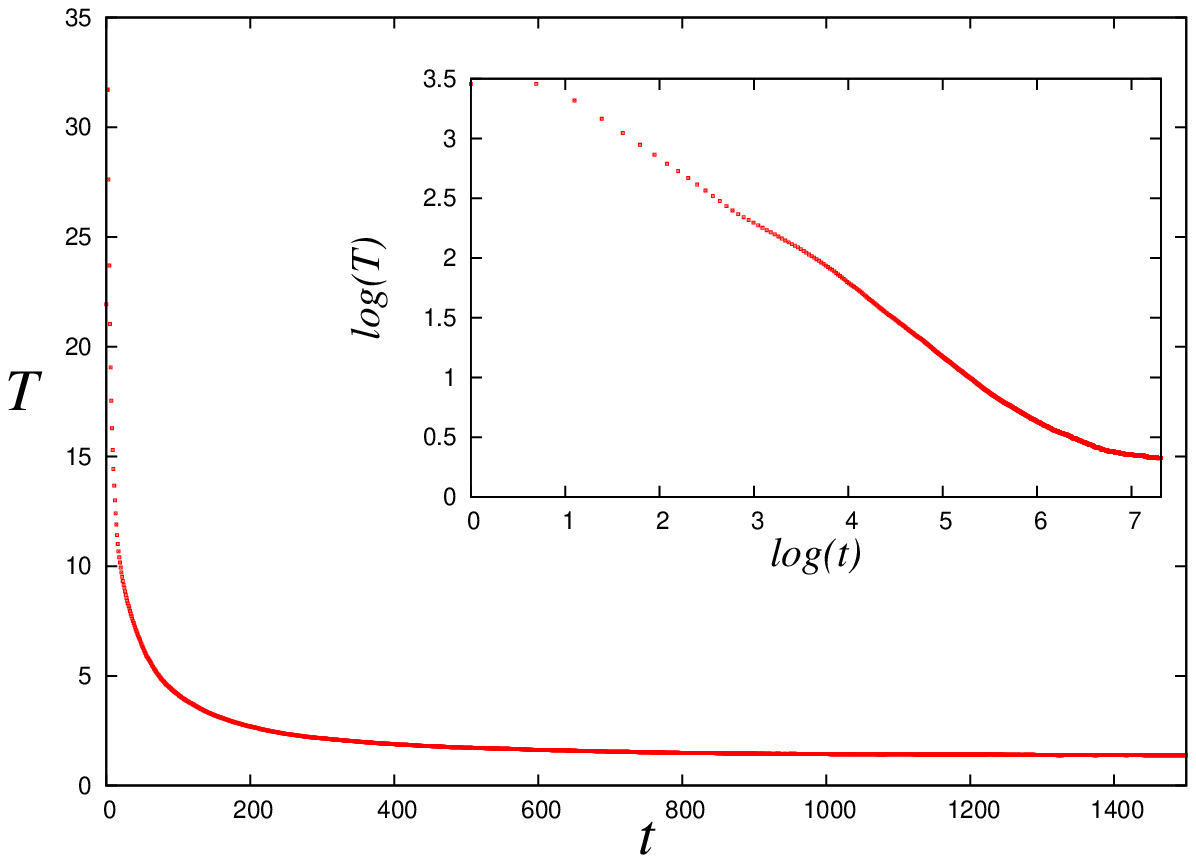}
\includegraphics[width=8cm]{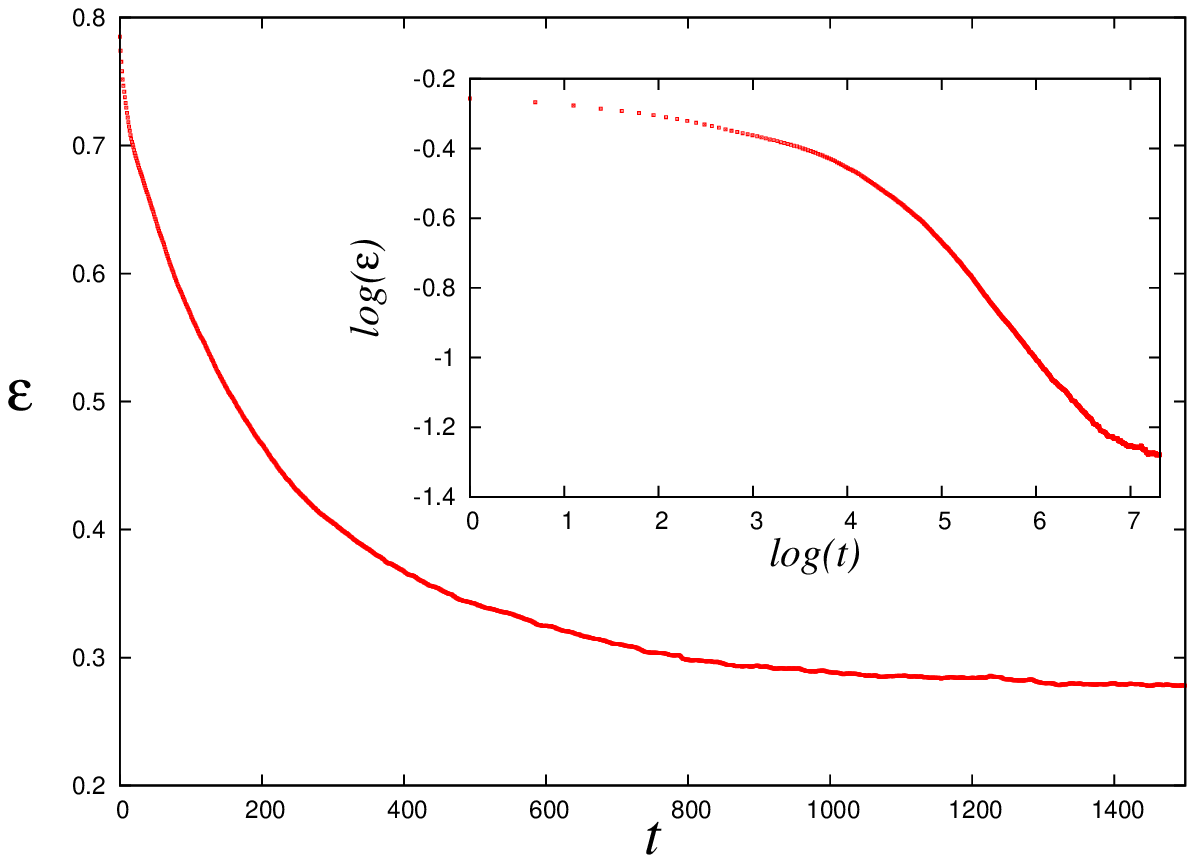}
\end{center}
\caption{\footnotesize 
Time evolution of 
the effective temperature (upper panel) and 
the residual energy defined by (\ref{eq:def_residual}) (lower panel) for the case of 
spin glass chain. We used a simple GA having  
$\sigma=2, p_{c}=0.1, p_{m}=0.001$. 
The inset stands for the asymptotic behaviour. }
  \label{fig:fg1D}
\end{figure}

%%%%%%%%%%%%%%%%%%%%%%%%%%%%%%%%%
We first show the time-evolution of 
effective temperature and the residual energy for the case of 
spin glass chain with parameter sets: 
$\sigma=2,p_c=0.1,p_m=0.001$
in Figure \ref{fig:fg1D}. 
From this figure, we find that 
the asymptotic behaviour of 
the effective temperature follows a power-law. 
This schedule is faster than the 
effective temperature scheduling 
for the optimal simulated annealing $\sim 1/\log (1+t)$, 
however, slower than the exponential decreasing. 
Thus, here we define the 
residual energy 
and its time-dependence 
as the difference between the 
lowest energy and current energy 
obtained by the GA dynamics. 
We find that 
the residual energy 
which is defined by 
%%%%
\begin{equation}
\varepsilon \equiv 
H(\bm{s})-
\min_{\bm{s}}H(\bm{s})
\label{eq:def_residual}
\end{equation}
%%%%%
also asymptotically goes to 
zero and 
it follows a power-law in the scaling regime $t \gg 1$. 
%%%%%%%%%%%%%%%%%%%%%%%%%%%
%%%%%%%%%%%%%%%%%%%%
\mbox{}

\begin{figure}[ht]
\begin{center}
\includegraphics[width=8cm]{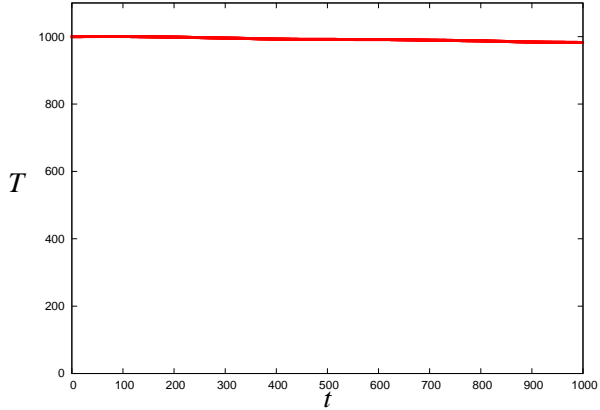}
\end{center}
\caption{\footnotesize 
Time evolution of 
the effective temperature for 
the case of spin glass chain by simple GA having 
$p_{c}=0.1, p_{m}=0.001$ and 
without any selection operation $\sigma=1$. }
\label{fig:fgNo}
\end{figure}

To investigate the 
effect of the selection operator on 
the GA dynamics, 
we carry out the same 
numerical experiments for 
the case of $\sigma=1$, 
namely, we investigate the 
time-evolution of 
the effective temperature for 
the GA without any effective selection 
(leading up to `random selection'). 
We plot the result in Figure \ref{fig:fgNo}. 
From this figure, we find that 
the effective temperature 
does  not decrease and remains the same value as 
the initial condition. 
This means that 
the behaviour of the effective temperature is 
strongly dependent on the selection．
%%%%%%%%%%%%%%%%%%%%%%%%%%%%%
%%%%%%%%%%%%%%%%%%%%%
%%%%%%%%%%%%%%%%%%%%%%%%%%%
%%%%%%%%%%%%%%%%%%%%%%%%%%%%%%%%%%%%%%%%%%%%%%

\begin{figure}[ht]
\begin{center}
\includegraphics[width=8cm]{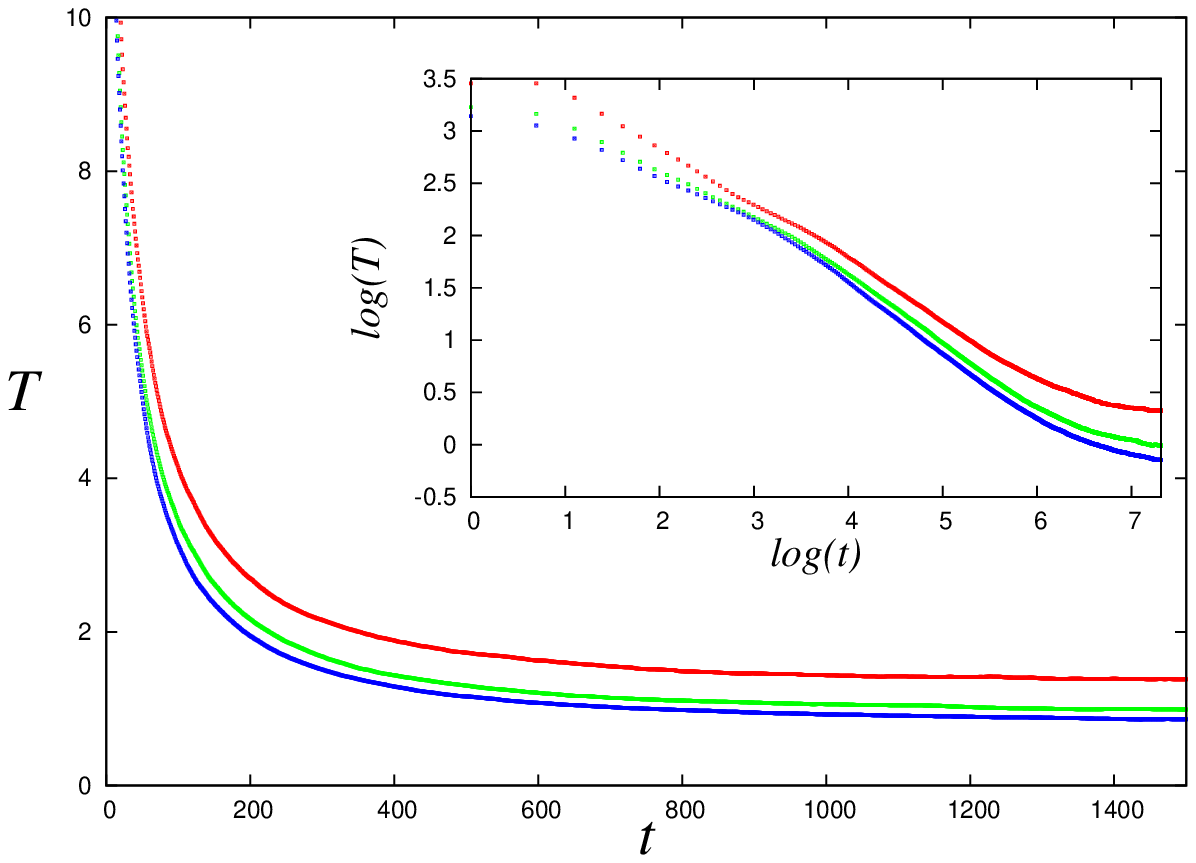}
\includegraphics[width=8cm]{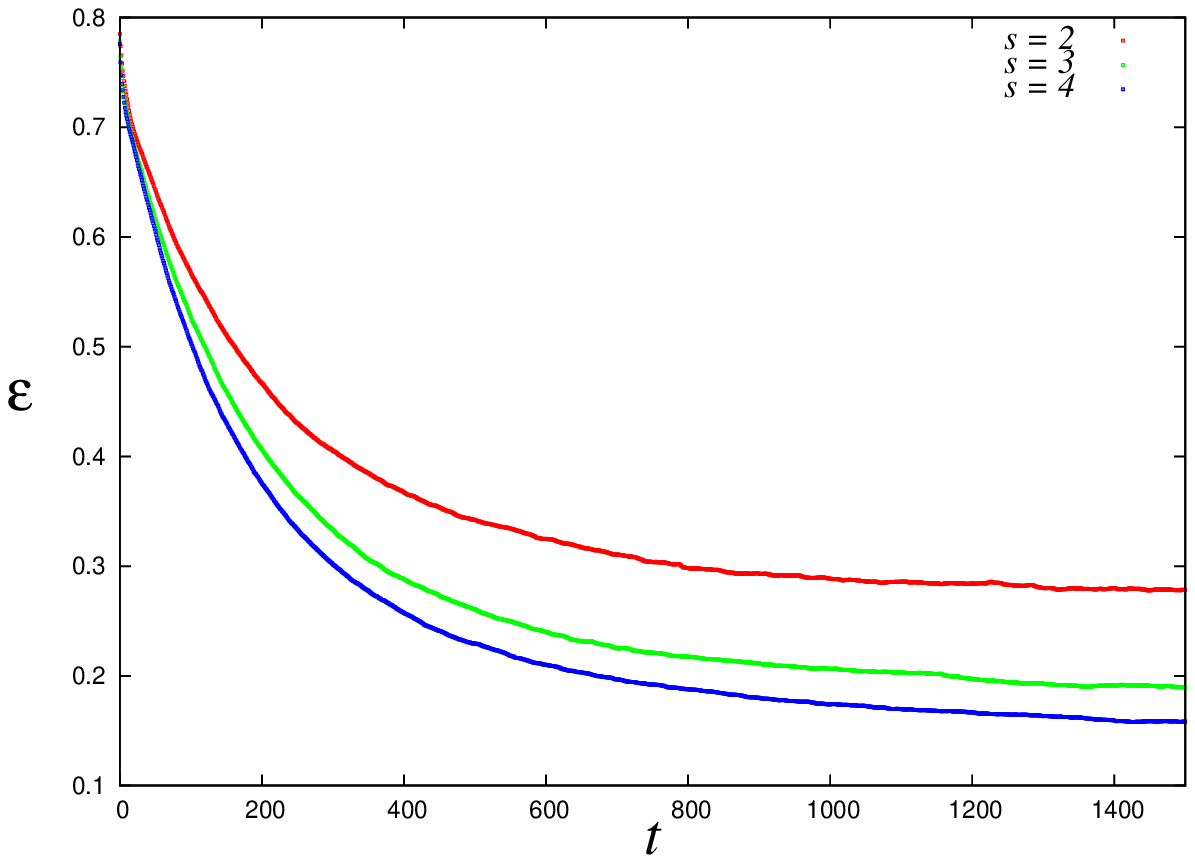}
 \end{center}
 \caption{\footnotesize 
Time evolution of 
effective temperature 
and residual energy 
defined by (\ref{eq:def_residual}) 
for the case of spin glass chain. 
We used a simple GA specified by $\sigma=2,3,4$ 
keeping $p_{c}=0.1$ and $p_{m}=0.001$. 
 }
   \label{fig:fgS}
\end{figure}

We next consider the relationship between the time-evolution 
of effective temperature, 
residual energy and 
the values of parameters for GA operations 
during the dynamics.  
We first fix $p_{c}=0.1, 
p_{m}=0.001$ and 
evaluate the result by changing 
the parameter $\sigma$ as $\sigma=2,3$ and $4$. 
The result is shown in Figure \ref{fig:fgS}. 
From these panels, we find that 
the speed of effective temperature decreasing  
for large $\sigma$ value 
is faster than the result for small $\sigma$ value. 
However, 
in the asymptotic regime, 
the behaviour of 
effective temperature is almost independent of 
the choice of $\sigma$ value．
%%%%%%%%%%%%%%%%%%%%%%%%%%%%%%%
%%%%%%%%%%%%%%%%%%%
\mbox{}

\begin{figure}[ht]
\begin{center}
 \includegraphics[width=8cm]{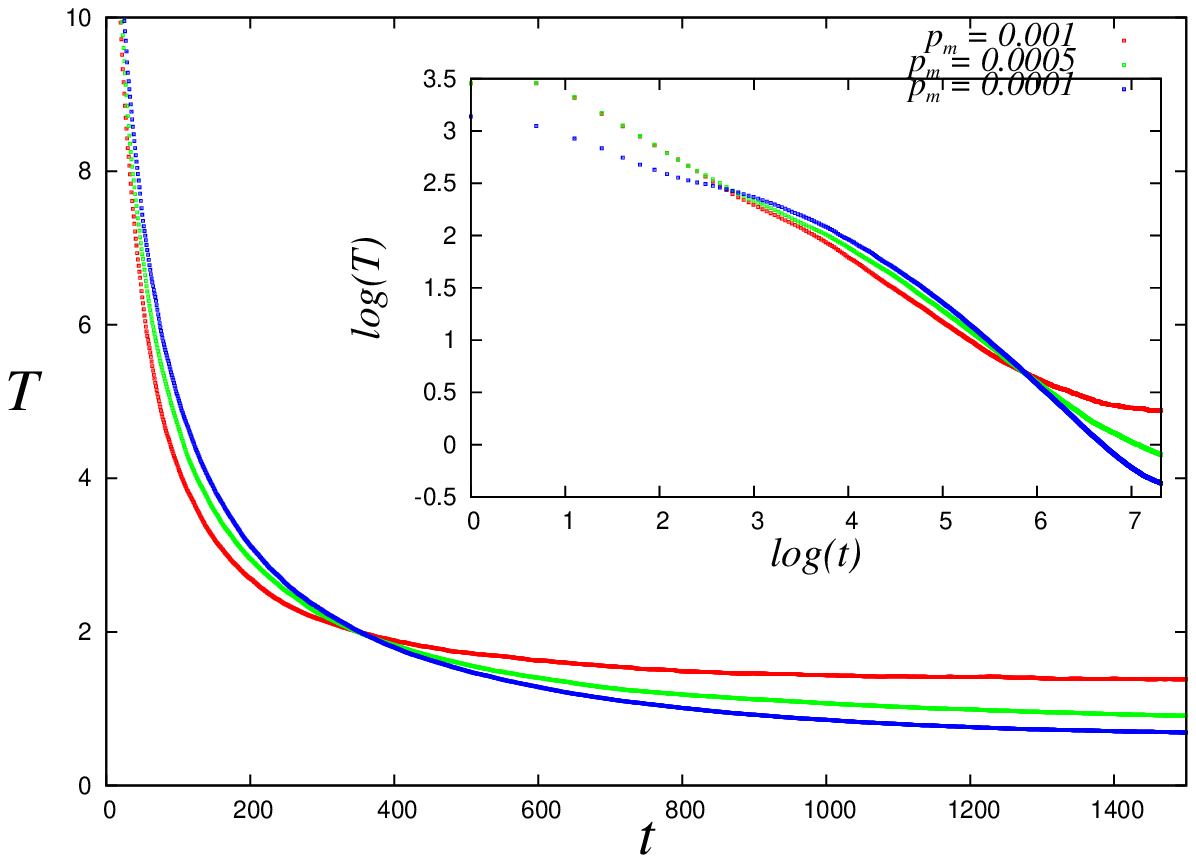}
\includegraphics[width=8cm]{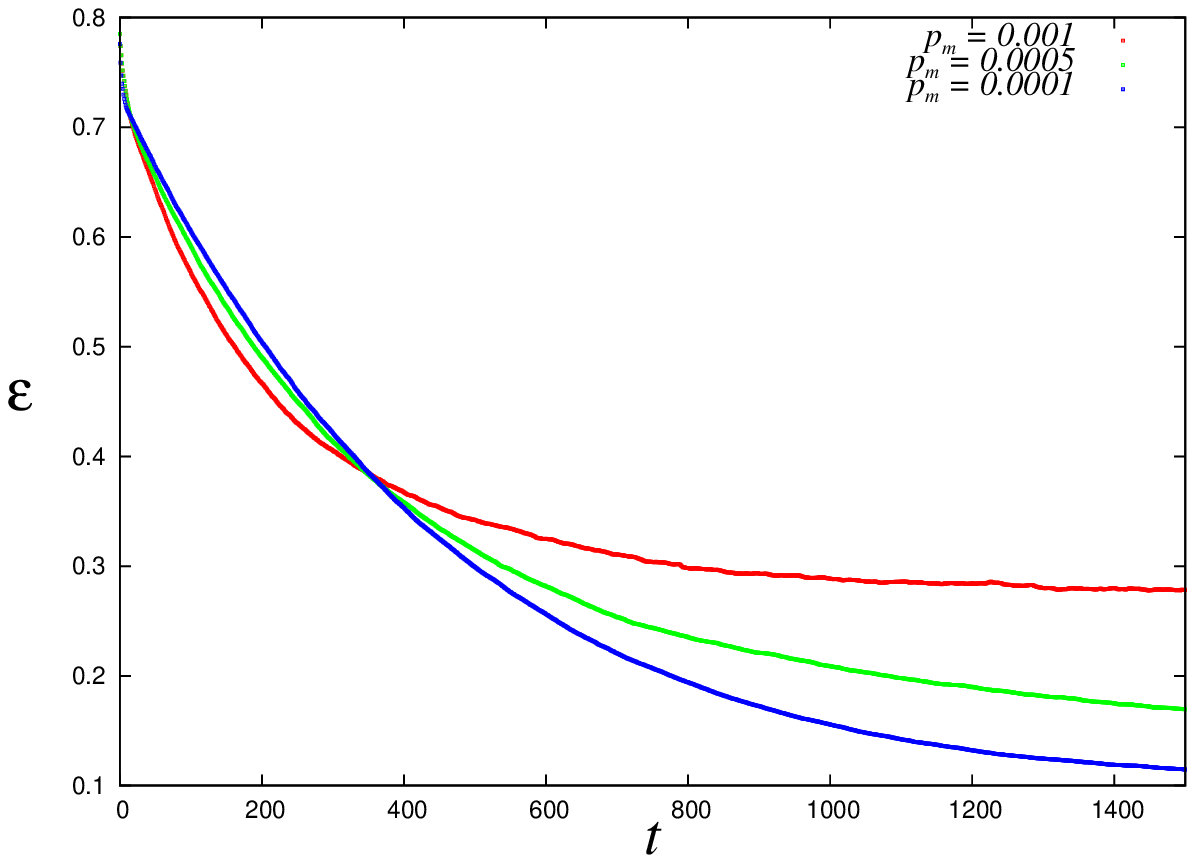}
  \end{center}
  \caption{\footnotesize 
Time evolution of the effective temperature (upper panel) and 
the residual energy defined by (\ref{eq:def_residual}) (lower panel)  
for 
the case of spin glass chain. We utilized a simple GA having 
$p_{m}=0.0005,0.001$ and $0.005$ keeping $p_{c}=0.1$ 
and $\sigma=2$. }
   \label{fig:fgM}
\end{figure}

We next consider the case of 
$p_{m}=0.0001,0.0005$ and $0.001$
keeping $\sigma=2$ and $p_{c}=0.1$. 
The result is shown in Figure 
\ref{fig:fgM}. 
From this figure, we confirm that 
the speed of convergence becomes very slow 
for both initial stage and asymptotic regime 
of the dynamics for  $p_{m}=0.001$. 
This result implies that  
`mixing' among the gene configurations is enhanced 
for large $p_{m}$ so as to  
prevent the Gibbs distribution from 
converging. 
On the other hand, 
for the case of 
$p_{m}=0.0005$ 
in the asymptotic regime, 
the speed of convergence is 
not so slow although 
the speed in 
the initial stage is actually slow.
We also find 
this result from the behaviour of the residual energy. 
The result for 
$p_{m}=0.0001$
gives the largest exponent 
$\xi$ of 
the asymptotic form
%%%%
\begin{equation}
 T (t) = 
t^{-\xi},\,\,\,
(t \gg 1)
\label{eq:scaling},
\end{equation}
%%%%% 
namely, the speed of convergence is the fastest among the 
three cases．
From the observation above, we find that mutation in a simple 
GA  makes the population diverse to 
prevent us from trapping in a local minima of 
energy function and one can 
enhanced the speed of convergence asymptotically 
by setting the parameter $p_{m}$ to an appropriate value. 
%%%%%%%%%%%%%%%%%%%%%%%%%%%%%%%
%%%%%%%%%%%%%%%%%%%
%%%%%%%%%%%%%%%%%%%%%%%%%%%%
\mbox{}

\begin{figure}[ht]
\begin{center}
 \includegraphics[width=8cm]{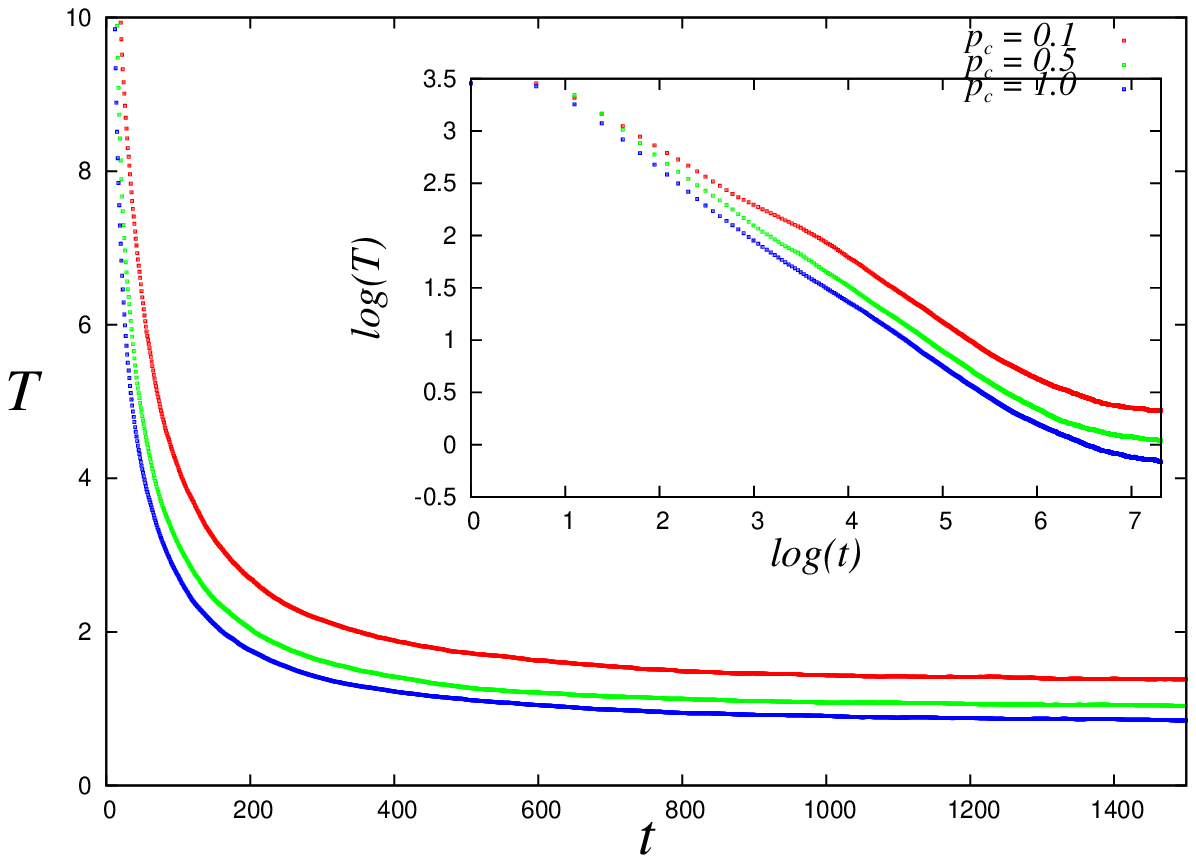}
 \includegraphics[width=8cm]{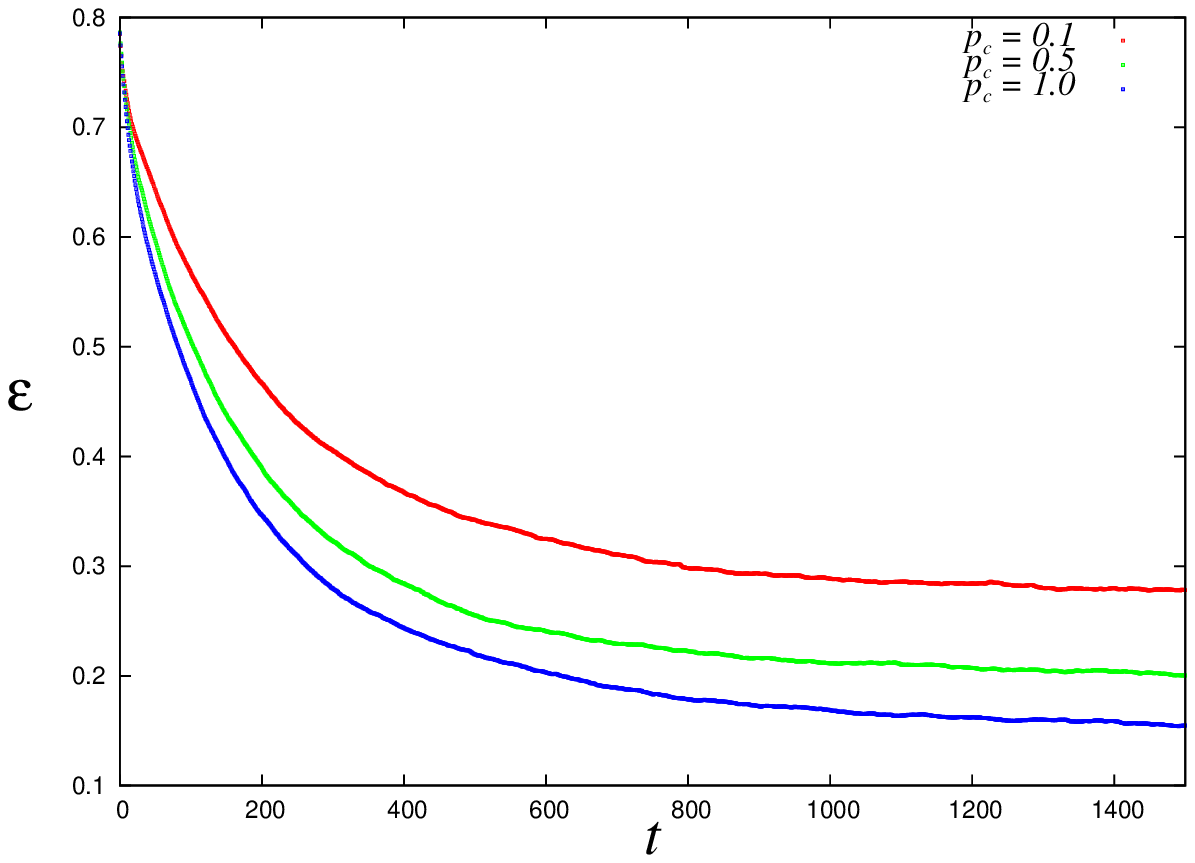}
  \end{center}
  \caption{\footnotesize 
Time evolution of the effective temperature (upper panel) and 
the residual energy 
defined by (\ref{eq:def_residual})
(lower panel)  
for the case of spin glass chain. We utilized a simple GA 
specified by $p_{c}=1, 0.5,0.1$ keeping $p_{m}=0.001$ 
and $\sigma=2$. 
 }
   \label{fig:fgC}
\end{figure}

Finally, we investigate the 
time-evolution of 
effective temperature 
for $p_{c}=1.0,0.5$ and $0.1$
keeping 
$\sigma=2$ and $p_{m}=0.001$. 
The result is shown in Figure \ref{fig:fgC}.
From this figure, we find that 
higher value of the crossover rate gives 
higher convergence of the effective temperature. 
Generally speaking, a crossover is one of the 
essential operators in GA to generate 
genes having good quality 
in terms of minimization of the cost. 
However, at the same time, 
one has some risks to destruct 
the good equality gene itself 
when we choose too large crossover rate. 
The cost function we deal with in 
this section is 
that of the spin glass chain and 
interactions among the spins 
exist only in the nearest neighboring spin pairs. 
This fact means that 
there is less possibility that the crossover deconstructs the 
fine genes in comparison with 
the case of the Sherrington-Kirkpatrick model 
which will be mentioned in the next subsection. 
Actually, for the case of 
$p_{c}=1.0$, the GA gives the best performance among the three cases 
$p_{c}=1.0,0.5$ and $0.1$. 
Nevertheless, in the asymptotic regime, 
the three cases gives almost the same performance. 
%%%%%%%%%%%%%%%%%%%%%%%%%%%%%%%
%%%%%%%%%%%%%%%%%%%
%%%%%%%%%%%%%%%%%%%
%%%%%%%%%%%%%%%%%%%%%%%%%%%%%%%%%%%%%%%%%%%%%%
\subsection{Sherrington-Kirkpatrick model}
%%%%%%%%%%%%%%%%%%%%%%%%%%%%%%%%%%%%%%%%%%%%%%
We next consider the case of the SK spin glass. 
In the SK model, it is difficult for us to 
obtain the exact 
lowest energy to evaluate the 
residual energy. 
Hence, here we investigate 
the time evolution of effective temperature 
and the average fitness which is 
defined as negative internal energy $-U=-H(\bm{s})$. 
%%%%%%%%%%%%%%%%%%%%%%%%%%%%%%%%%%%%%%%%
%%%%%%%%%%%%%%%%%%%%%%%%%%%%%%%%%%%%%%%%%%%%%%

\begin{figure}[ht]
\begin{center}
   \includegraphics[width=8cm]{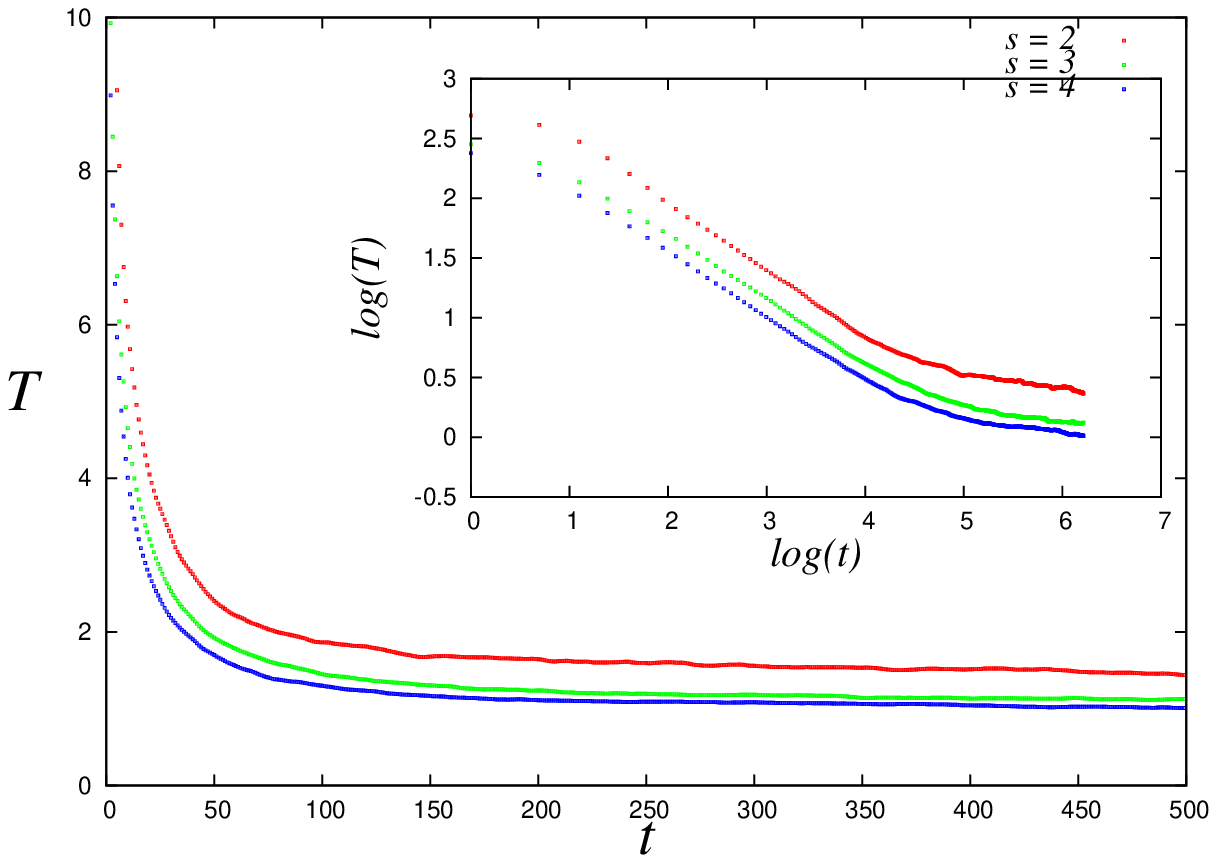}
 \includegraphics[width=8cm]{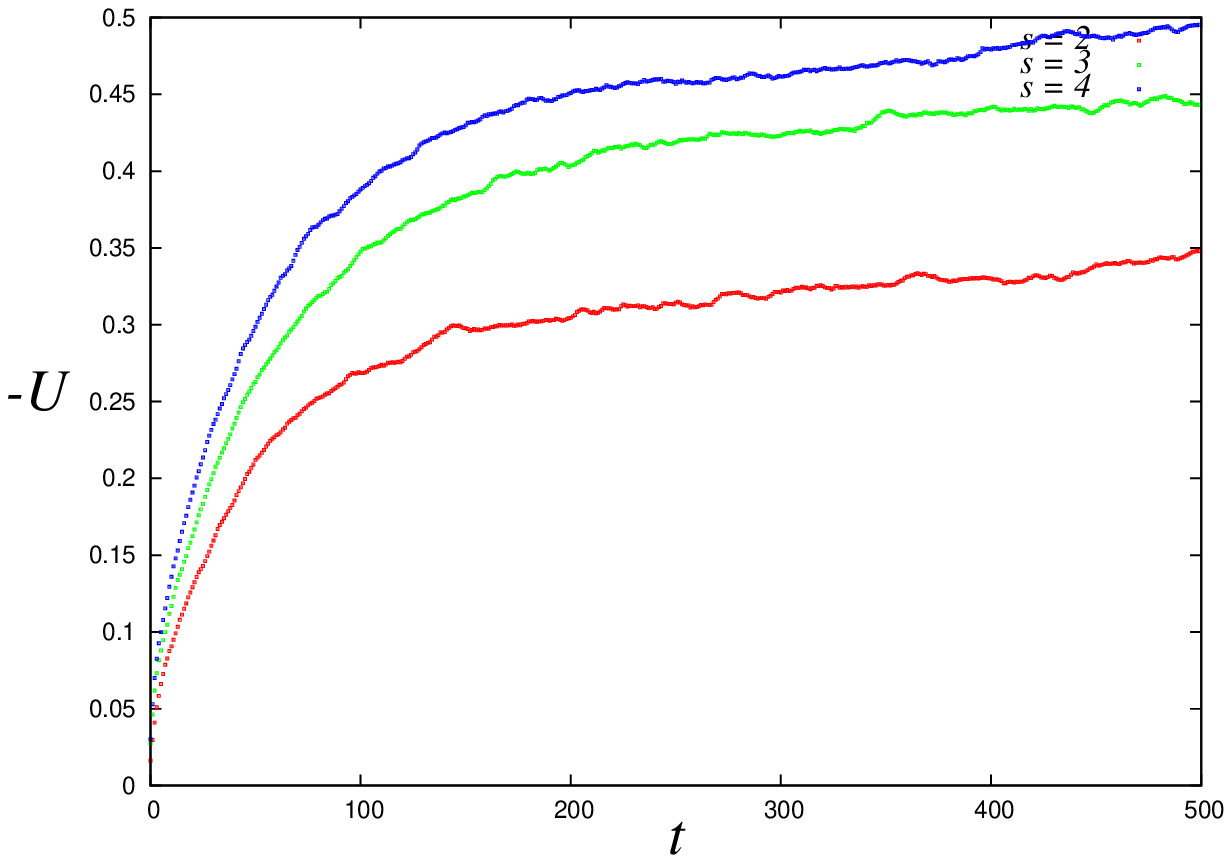}
  \end{center}
  \caption{\footnotesize
Time evolution of 
the effective temperature (upper panel)  
and the averaged  fitness which is 
defined as negative internal energy $-U$ (lower panel) 
for the case of SK model. We used a simple GA 
having $\sigma=2,3$ and $4$
keeping $p_{m}=0.005$ and 
$p_{c}=0.05$. 
In the asymptotic regime $t \gg 1$ 
of time-evolution of temperature, 
`crossover phenomena' are observed. 
Namely, the power-law exponent 
$\xi$ changes to the different value at intermediate 
time scale $\log t \sim 5$. 
}
   \label{fig:fgSSK}
\end{figure}

As we discussed in the previous subsection, 
we first investigate the time-evolution of 
these two physical quantities 
for the case of 
$\sigma=2,3$ and $4$ keeping 
$p_{c}=0.05, p_{m}=0.005$. 
We show the result in Figure \ref{fig:fgSSK}.  
From these panels, we find that 
the asymptotic performance 
through the effective temperature does not change 
even if we increases the $\sigma$ value. 
However, it should be noticed that 
some `crossover phenomena'  
takes place in some generation (time) regime. 
Namely, in this generation regime, 
the exponent $\xi$ in a power-law 
changes to the different exponent $\xi^{'}$ ($>\xi$).  
On the other hand, 
at the beginning of the evolution, 
the average fitness value 
increases 
as the $\sigma$ value increases. 
%%%%%%%%%%%%%%%%%%%%%%%%%%%%%%%
%%%%%%%%%%%%%%%%%%%
%%%%%%%%%%%%%%%%%%%%%%%
\mbox{}

\begin{figure}[ht]
\begin{center}
   \includegraphics[width=8cm]{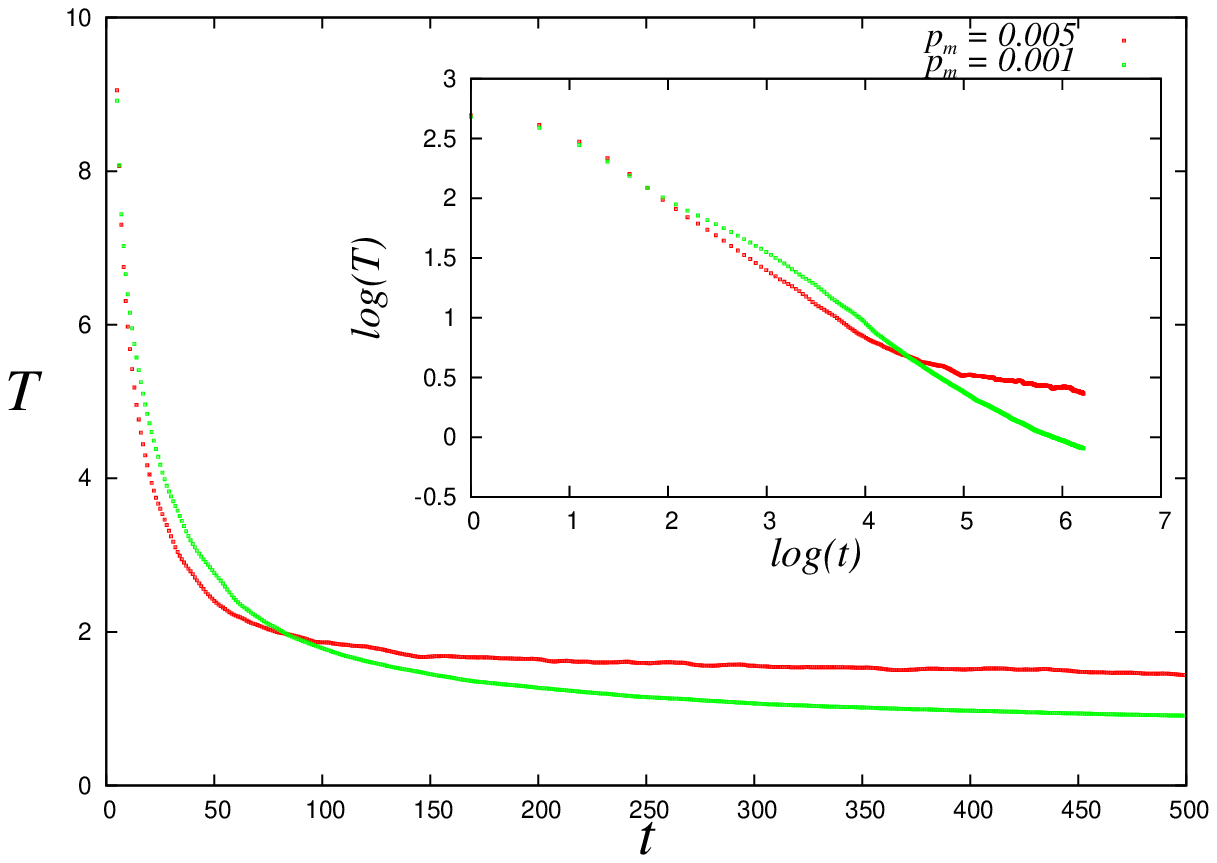}
 \includegraphics[width=8cm]{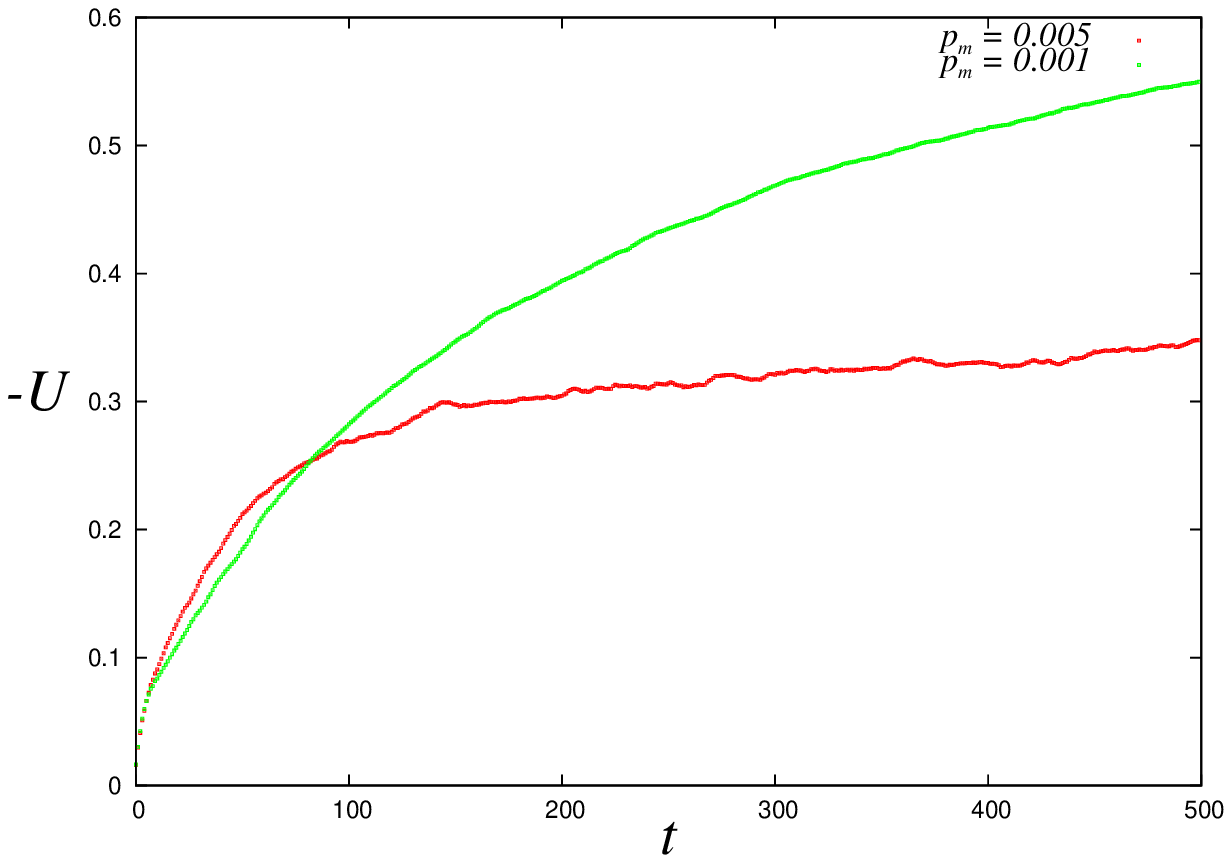}
  \end{center}
  \caption{\footnotesize
Time evolution of 
the effective temperature (upper panel)  
and the averaged  fitness which is 
defined as negative internal energy $-U$ (lower panel) 
for the case of SK model. We used a simple GA having 
$p_{m}=0.005,0.001$ keeping $p_{c}=0.05$ and 
$\sigma=2$. 
In the asymptotic regime $t \gg 1$ 
of time-evolution of temperature for $p_{m}=0.005$, 
`crossover phenomena' are observed. 
Namely, the power-law exponent 
$\xi$ changes to the different value at intermediate 
time scale $\log t \sim 5$. 
}
   \label{fig:fgMSK}
\end{figure}

We next consider the case of 
$p_{m}=0.005,0.001$
keeping $s=2$ and $p_{c}=0.05$. 
The results are shown in 
Figure \ref{fig:fgMSK}. 
From this figure, we confirm that 
the speed of convergence  
for the case of $p_{m}=0.005$ 
slows down in the asymptotic regime
whereas the speed for the case of 
$p_{m}=0.001$ remains. 
The same behaviour as time evolution of 
the effective temperature is observed in the lower panel of 
Figure \ref{fig:fgMSK}.
%%%%%%%%%%%%%%
%%%%%%%%%%%%%%%%%%%%%%%%%%%%%%%
%%%%%%%%%%%%%%%%%%%
%%%%%%%%%%%%%%%%%%%%%%
\mbox{}

\begin{figure}[ht]
\begin{center}
 \includegraphics[width=8cm]{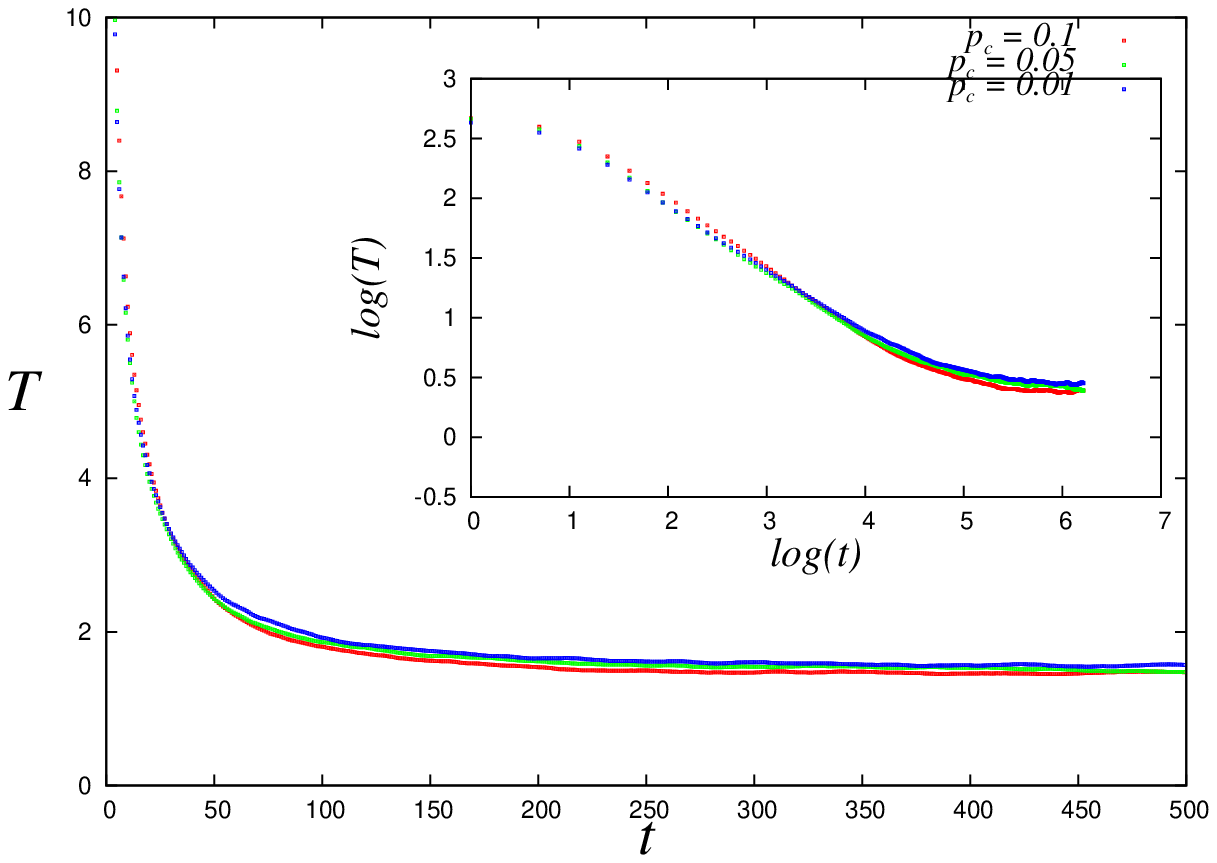}
 \includegraphics[width=8cm]{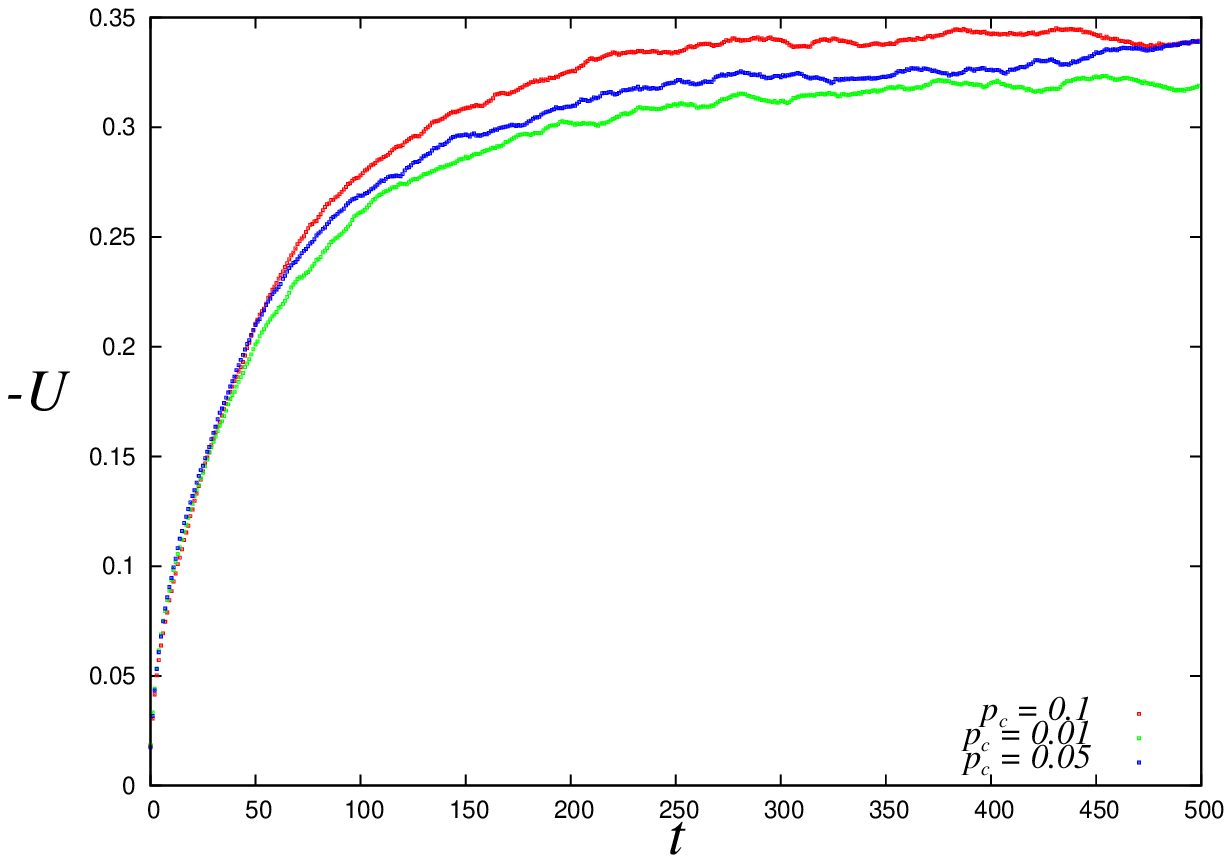}
\end{center}
\caption{\footnotesize
Time evolution of 
the effective temperature (upper panel)  
and the averaged  fitness which is 
defined as negative internal energy $-U$ (lower panel)  
for the case of SK model. A simple GA having 
$p_{c}=0.1,0.05, 0.01$ keeping $p_{m}=0.005$ and 
$\sigma=2$ is utilized. }
\label{fig:fgCSK}
\end{figure}

Finally, we shall show the results for 
$p_{c}=0.1,0.05$ and $0.01$ keeping $\sigma=2$ and $p_{m}=0.005$ in 
Figure \ref{fig:fgCSK}. 
As the SK model is defined on a complete graph and 
all spins are connected, 
the crossover operation might destroy the gene configurations 
having relatively high fitness values. 
However, 
from the results shown in 
this figure, 
the average fitness value increases as the 
$p_{c}$ increases although the effective temperature 
does not change so much. 
%%%%%%%%%%%%%%%%%
%%%%%%%%%%%%%%%%%%%%%%%%%%%%%%%
%%%%%%%%%%%%%%%%%%%%%%%%%%%%%%%%%%%%%%%%%
\section{Concluding remarks}
%%%%%%%%%%%%%%%%%%%%%%%%%%%%%%%%%%%%%%%%%
In this paper, we introduced a learning algorithm 
of Gibbs distributions 
from training sets which are 
gene strings generated 
by GA to figure out 
the statistical properties of GA from 
the view point of thermodynamics. 
A procedure of 
average-case performance evaluation for genetic algorithms 
was numerically examined. 
The formulation was applied to 
the solvable probabilistic models 
having multi-valley energy landscapes, 
namely, the spin glass chain and 
the Sherrington-Kirkpatrick model. 
By using computer simulations, we discussed the asymptotic 
behaviour of the effective 
temperature scheduling and the residual energy induced by the GA dynamics. 

Both effective temperature and residual energy 
show power-law behaviors given 
by (\ref{eq:scaling}), namely, 
$\beta_{t} = t^{\xi}$ for $t \gg 1$. 
In section 2, we showed that  
a Gibbs distribution with 
$\beta_{t}=t$ yields  
the Holland's condition (\ref{eq:holland1}). 
Hence, it might be worth while for us to 
check to what extent the condition 
is modified for 
the Gibbs distribution with 
$\beta_{t}=t^{\xi}$. 
For this case, a Gibbs distribution with respect to 
the gene configuration having 
the fitness $g(i)$ is written as 
$p_{i}(t)  =  
{\exp}[t^{\xi}g(i)]/
\sum_{j \in \mathcal{J}}
{\exp}[t^{\xi}g(i)]$. 
Taking the derivative of the above equation 
with respect to 
$t$, we have 
$dp_{i}(t)/dt =  
\xi t^{\xi-1}
\{
p_{i}(t)g(i)
-p_{i}(t)
\sum_{j \in \mathcal{J}}
g(i)p_{j}(t)\}$. 
%%%% 
Hence, 
by substituting this result into 
the equation obtained by 
taking the derivative of 
the probability $P(\mathcal{H},t)$ that a schema 
$\mathcal{H}$ appears, namely, 
$P(\mathcal{H},t)=\sum_{i \in \mathcal{H}}p_{i}(t)$ 
with respect to 
$t$, we obtained the modified 
Holland's condition as 
${d P(\mathcal{H},t)}/{dt}  =  
\xi t^{\xi -1}
\{
f(\mathcal{H},t)-P(\mathcal{H},t)f(\mathcal{J},t)\}$.
%%%%% 
From this condition, we find that 
the temporal difference of the probability 
that the schema $\mathcal{H}$ appears 
increases by $\xi t^{\xi-1}$ 
for $\beta_{t}=\xi t^{\xi-1}$. 
More generally, we conclude that 
the temporal difference 
increases by $d\beta_{t}/dt$ for 
$\beta_{t}$.
 
Although we dealt with 
the average-case performance 
evaluation just for a simple GA, 
our general procedure given in this paper is apparently  
applicable to the other sophisticated GAs based on 
any population dynamics. Moreover, 
one can generalize the Gibbs form to be trained 
by Boltzmann-machine-type learning equation 
so as to include the so-called Tsallis distribution, 
which is specified by $\beta$ and $q$,  as 
a special case \cite{Nishimori}. 
%%%%%%%%%%%%%%%%%%%%%%%%%%%%%%%%%%%%%%%%%
\section*{Acknowledgments}
%%%%%%%%%%%%%%%%%%%%%%%%%%%%%%%%%%%%%%%%
We were financially supported by Grant-in-Aid Scientific Research on
Priority Areas {\it `Deepening and Expansion of Statistical Mechanical Informatics
(DEX-SMI)'} of the MEXT No. 18079001. 
One of the authors (JI) was financially supported by 
INSA (Indian National Science Academy) -  JSPS 
(Japan Society of Promotion of Science)  Bilateral Exchange Programme. 
He also thanks Saha Institute of Nuclear Physics for their warm hospitality during 
his stay in India. 
%%%%%%%%%%%%%%%%%%%%%%%%%%%%%%%%%%%%%%%%%
\renewcommand{\baselinestretch}{0.98}
\bibliographystyle{apalike}
{\small
\bibliography{example}}

\begin{thebibliography}{}

\bibitem[Baluja, 1994]{Baluja}
Baluja, S. (1994).
\newblock Population-based incremental learning: A method for integrating
  genetic search based function optimization and competitive learning.
\newblock {\em Technical Report, School of Computer Science, Carnegie Mellon
  University}, CMU-CS-94:163.

\bibitem[E.Goldberg, 1989]{Goldberg}
E.Goldberg, D. (1989).
\newblock {\em Genetic Algorithms in Search, Optimization and Machine
  Learninig}.
\newblock Addison-Wesley.

\bibitem[Geman and Geman, 1984]{Geman}
Geman, S. and Geman, D. (1984).
\newblock Stochastic relaxation, gibbs distributions, and the bayesian
  restoration of images.
\newblock {\em IEEE Trans. on Pattern Analysis and Machine Intelligence},
  PAMI-6:721--741.

\bibitem[H.Chen and K.Ma, 1982]{Chen}
H.Chen, H. and K.Ma, S. (1982).
\newblock Low-temperature behaviour of a one-dimensional ising model.
\newblock {\em Journal of Statsitcal Physics}, 29:717--746.

\bibitem[H.Holland, 1975]{Holland}
H.Holland, J. (1975).
\newblock {\em Adaptation in natural and artificial systems}.
\newblock The University of Michigan Press.

\bibitem[Kirkpatrick et~al., 1983]{Kirkpatrick}
Kirkpatrick, S., D.Galatt, C., and P.Vecchi, M. (1983).
\newblock Optimization by simulated annealing.
\newblock {\em Science}, 220:671--680.

\bibitem[Li, 1981]{Li}
Li, T. (1981).
\newblock Structure of metastable states in a random ising chain.
\newblock {\em Physical Review B}, 24:6579--6587.

\bibitem[L.Shapiro, 2005]{Shapiro}
L.Shapiro, J. (2005).
\newblock Drift and scaling in estimation of distribution algorithms.
\newblock {\em Evolutionary Computation}, 13.

\bibitem[L.Shapiro, 2006]{Shapiro2}
L.Shapiro, J. (2006).
\newblock Diversity loss in general estimation of distibution algorithms.
\newblock {\em Lecture Notes in Computer Science}, 4193.

\bibitem[Mezard and Parisi, 1986]{Mezard}
Mezard, M. and Parisi, G. (1986).
\newblock A replica analysis of the travelling salesman problem.
\newblock {\em Journal de Physique}, 47:1285--1296.

\bibitem[Mezard et~al., 1987]{SpinGlass}
Mezard, M., Parisi, G., and Virasoro, M. (1987).
\newblock {\em Spin Glass Theory and Beyond}.
\newblock World Scientific, Singapore.

\bibitem[Monasson et~al., 1999]{Monasson}
Monasson, R., Zecchina, R., Kirkpatrick, S., Selman, B., and Troyansky, L.
  (1999).
\newblock Determining computational complexity from characteristic `phase
  transitions'.
\newblock {\em Nature}, 400:133--137.

\bibitem[Nishimori and Inoue, 1998]{Nishimori}
Nishimori, H. and Inoue, J. (1998).
\newblock Convergence of simulated annealing using the generalized transition
  probability.
\newblock {\em Journal of Physics A: Mathematical and Genetal}, 47:5561--5672.

\bibitem[Pelikan et~al., 1999]{Pelikan1}
Pelikan, M., E.Goldberg, D., and E.Cantu-Paz, E. (1999).
\newblock Boa: The bayesian optimization algorithm.
\newblock In {\em Proceedings of GECCO-99}.

\bibitem[Pelikan et~al., 2000]{Pelikan2}
Pelikan, M., E.Goldberg, D., and E.Cantu-Paz, E. (2000).
\newblock Bayesian optimization algorithm, population sizing, and time to
  convergence.
\newblock {\em University of Illinois at Urbana-Champaign, Illinois Genetic
  Algorithms Laboratory, Urbana, IL, IlliGAL Report}, No. 2000002.

\bibitem[Pelikan et~al., 2002]{Pelikan3}
Pelikan, M., E.Goldberg, D., and G.Lobo, F. (2002).
\newblock Survey of optimization by building and using probabilistic models.
\newblock {\em Computational Optimization and Applications}, 21:5--20.

\bibitem[Prugel-Bennett and L.Shapiro, 1994]{Bennett1}
Prugel-Bennett, A. and L.Shapiro, J. (1994).
\newblock An analysis of genetic algorithms using statistical mechanics.
\newblock {\em Physical Review Letters}, 72:1305--1309.

\bibitem[Prugel-Bennett and L.Shapiro, 1997]{Bennett2}
Prugel-Bennett, A. and L.Shapiro, J. (1997).
\newblock The dynamics of a genetic algorithm for simple ising systems.
\newblock {\em Physica D}, 104:75--114.

\bibitem[S.Correa and L.Shapiro, 2006]{Shapiro3}
S.Correa, E. and L.Shapiro, J. (2006).
\newblock Model complexity vs. performance in the bayesian optimization
  algorithm.
\newblock {\em Lecture Notes in Computer Science}, 4193.

\bibitem[Sherrington and Kirkpatrick, 1975]{SK}
Sherrington, D. and Kirkpatrick, S. (1975).
\newblock Solvable model of spin-glass.
\newblock {\em Physical Review Letters}, 35:1792--1796.

\bibitem[Suzuki, 1995]{Suzuki}
Suzuki, J. (1995).
\newblock Markov chain analysis of simple genetic algorithm.
\newblock {\em IEEE Trans. on System, Man and Cybernatics}, 25:655--659.

\bibitem[Suzuki, 1998]{Suzuki2}
Suzuki, J. (1998).
\newblock A further result on the markov chain model od gas and their sa-like
  strategy.
\newblock {\em IEEE Trans. on System, Man and Cybernatics}, 25:95--102.

\bibitem[Suzuki, 2005]{Suzuki3}
Suzuki, J. (2005).
\newblock Statsitical physics approach to genetic algorithm (in japanese).
\newblock In {\em Proceedings of Computational Intelligence Seminar (8th
  November, 2005, Waseda University, Tokyo Japan)}.

\bibitem[Talagrand, 2003]{Talagrand}
Talagrand, M. (2003).
\newblock {\em Spin Glasses: A Challenge for Mathematicians}.
\newblock Springer.

\end{thebibliography}

\end{document}